\documentclass{article}

\usepackage[utf8]{inputenc}

\usepackage{longtable}
\usepackage{amsmath,amssymb,amsfonts}
\usepackage{algorithmic}
\usepackage{graphicx}
\usepackage{textcomp}
\usepackage{xcolor}
\usepackage{hyperref}
\usepackage{nameref}
\usepackage{multirow}
\usepackage{multicol}
\usepackage{caption}
\usepackage{subcaption}
\usepackage[a4paper,tmargin=30mm,bmargin=30mm,lmargin=15mm,rmargin=15mm,headsep=20mm]{geometry}
\usepackage{float}

\usepackage[acronym, shortcuts]{glossaries}

\makeglossaries

\newacronym{1nn}{1-NN}{1-Nearest Neighbor classifier}
\newacronym{acgan}{ACGAN}{Auxiliary Classifier GAN}
\newacronym{adain}{ADAIN}{Adaptive Instance Normalization}
\newacronym{ai}{AI}{Artificial Intelligence}
\newacronym{amt}{AMT}{Amazon Mechanical Turk}
\newacronym{ann}{ANN}{Artificial Neural Network}
\newacronym{aem}{AEM}{Attention Embedding Module}
\newacronym{c2st}{C2ST}{Classifier Two-sample Test}
\newacronym{c2stnn}{C2ST-NN}{C2ST-Neural Network}
\newacronym{cegan}{CEGAN}{Classification Enhancement GAN}
\newacronym{cgan}{CGAN}{Conditional GAN}
\newacronym{ce}{CE}{Conditional Embedding}
\newacronym{cmb}{CM-B}{Conditional Manipulating Block}
\newacronym{cnn}{CNN}{Convolutional Neural Network}
\newacronym{coco}{COCO}{Common Objects in Context}
\newacronym{csgan}{CSGAN}{Cyclic-Synthesized GAN}
\newacronym{cub}{CUB}{Caltech-UCSD Birds}
\newacronym{cuhk}{CUHK}{Chinese University of Hong Kong}
\newacronym{cyclegan}{CycleGAN}{Cycle-Consistent GAN}
\newacronym{dag}{DAG}{Data Augmentation Optimized for GAN}
\newacronym{dcgan}{DCGAN}{Deep Convolutional GAN}
\newacronym{dfgan}{DF-GAN}{Deep Fusion GAN}
\newacronym{dggan}{DGGAN}{Dynamically Grown GAN}
\newacronym{discogan}{DiscoGAN}{Learning to Discover Cross-Domain Relations with GANs}
\newacronym{dmgan}{DM-GAN}{Dynamic Memory GAN}
\newacronym{dualattentionalgan}{DualAttn-GAN}{Dual Attentional GAN}
\newacronym{dualgan}{DualGAN}{Unsupervised Dual Learning for Image-to-Image Translation}
\newacronym{eqgan}{EQGAN}{Entangling Quantum GAN}
\newacronym{f0}{F0}{Fundamental Frequency}
\newacronym{ffhq}{FFHQ}{Flickr-Faces-HQ}
\newacronym{fid}{FID}{Fréchet inception distance}
\newacronym{fldisco}{FL-DISCO}{Federated Generative Adversarial Network for Graph-based Molecule Drug Discovery}
\newacronym{gan}{GAN}{Generative Adversarial Network}
\newacronym{ganintcls}{GAN-INT-CLS}{Generative Adversarial Text to Image Synthesis}
\newacronym{gawwn}{GAWWN}{Generative Adversarial What-Where Network}
\newacronym{gpt3}{GPT-3}{Generative Pre-trained Transformer 3}
\newacronym{hr}{HR}{Human rank}
\newacronym{infogan}{InfoGAN}{Interpretable Representation Learning by Information Maximizing GANs}
\newacronym{ifd}{IFD}{Information Flow Diagram}
\newacronym{is}{IS}{Inception Score}
\newacronym{ldcgan}{LD-CGAN}{Lightweight Dynamic Conditional GAN}
\newacronym{leakyrelu}{LeakyReLU}{Leaky Rectified Linear Unit}
\newacronym{lpips}{LPIPS}{Learned Perceptual Image Patch Similarity}
\newacronym{ls-gan}{LS-GAN}{Loss-Sensitive GAN}
\newacronym{lsgan}{LSGAN}{Least Square GAN}
\newacronym{lstm}{LSTM}{Long Short-Term Memory}
\newacronym{ms}{MS}{Mode Score}
\newacronym{miegan}{MIEGAN}{Mobile Image Enhancement GAN}
\newacronym{mis}{m-IS}{modified-Inception Score}
\newacronym{missgan}{MISS GAN}{Multi-IlluStrator Style GAN}
\newacronym{mmd}{MMD}{Maximum Mean Discrepancy}
\newacronym{mwgan}{MWGAN}{Multi-marginal Wasserstein GAN}
\newacronym{mse}{MSE}{Mean Squared Error}
\newacronym{msssim}{MS-SSIM}{Multi-scale structural similarity for image quality}
\newacronym{mvc}{MVC}{Model-View-Controller}
\newacronym{ne}{NE}{Nash Equilibrium}
\newacronym{nlp}{NLP}{Natural Language Processing}
\newacronym{ntm}{NTM}{Neural Turing Machine}
\newacronym{parb}{PAR-B}{Pyramid Attention Refine Block}
\newacronym{pix2pix}{Pix2Pix}{Image-to-Image Translation with Conditional Adversarial Nets}
\newacronym{progan}{ProGAN}{Progressive Growing of GANs}
\newacronym{ps2man}{PS2MAN}{Photo-Sketch Synthesis using Multi-Adversarial Networks}
\newacronym{psnr}{PSNR}{Peak Signal to Noise Ratio}
\newacronym{qugan}{QuGAN}{A GAN Through Quantum States}
\newacronym{qram}{QRAM}{Quantum Random Access Memory}
\newacronym{relu}{ReLU}{Rectified Linear Unit}
\newacronym{rl}{RL}{Reinforcement Learning}
\newacronym{rnn}{RNN}{Recurrent Neural Network}
\newacronym{sagan}{SAGAN}{Self Attention GAN}
\newacronym{seqgan}{SeqGAN}{Sequence GAN}
\newacronym{smiles}{SMILES}{Simplified Molecular Input Line Entry Specification}
\newacronym{sngan}{SN-GAN}{Spectral Normalization for GANs}
\newacronym{srcagan}{srcaGAN}{Super Resolution Channel Attention GAN}
\newacronym{srgan}{SRGAN}{Super Resolution GAN}
\newacronym{ssdgan}{SSD-GAN}{Measuring the Realness in the Spatial and Spectral Domains}
\newacronym{ssgan}{SSGAN}{Self-supervised GAN}
\newacronym{ssim}{SSIM}{Structural Similarity Index Measure}
\newacronym{stackgan}{StackGAN}{Stacked GANs}
\newacronym{stackgan++}{StackGAN-v2}{StackGAN++}
\newacronym{stylegan}{StyleGAN}{A Style-Based Generator Architecture for Generative Adversarial Networks}
\newacronym{tam}{TAM}{Textual Attention Module}
\newacronym{textgan}{textGAN}{Text GAN}
\newacronym{tfd}{TFD}{Toronto Face Database}
\newacronym{ugan}{UGAN}{Unrolled GAN}
\newacronym{vam}{VAM}{Visual Attention Module}
\newacronym{wgan}{WGAN}{Wasserstein GAN}
\newacronym{wgangp}{WGAN-GP}{Gradient Penalty WGAN}
\newacronym{wsrgan}{WSRGAN}{Weighted SRGAN}
\newacronym{ylgan}{YLGAN}{Your Local GAN}

\usepackage{xspace}
\newcommand*{\eg}{e.g.\@\xspace}
\newcommand*{\ie}{i.e.\@\xspace}

\begin{document}

\title{A survey on GANs for computer vision: Recent research, analysis and taxonomy}

\author{Guillermo Iglesias
\thanks{Departamento de Sistemas Inform\'aticos, Universidad Polit\'ecnica de Madrid. ETSISI, Campus Sur, C/Alan Turing, s/n, 28031, Madrid, Spain}
\\{guillermo.iglesias@upm.es}
\and
Edgar Talavera
\footnotemark[1]
\\{e.talavera@upm.es (Corresponding author)}
\and
Alberto Díaz-Álvarez
\footnotemark[1]
\\{alberto.diaz@upm.es}}

\date{}

\maketitle

\begin{abstract}
    In the last few years, there have been several revolutions in the field of deep learning, mainly headlined by the large impact of \glspl{gan}. \glspl{gan} not only provide an unique architecture when defining their models, but also generate incredible results which have had a direct impact on society. Due to the significant improvements and new areas of research that \glspl{gan} have brought, the community is constantly coming up with new researches that make it almost impossible to keep up with the times. Our survey aims to provide a general overview of \glspl{gan}, showing the latest architectures, optimizations of the loss functions, validation metrics and application areas of the most widely recognized variants. The efficiency of the different variants of the model architecture will be evaluated, as well as showing the best application area; as a vital part of the process, the different metrics for evaluating the performance of \glspl{gan} and the frequently used loss functions will be analyzed. The final objective of this survey is to provide a summary of the evolution and performance of the \glspl{gan} which are having better results to guide future researchers in the field.
\end{abstract}

\section{Introduction}

\acrfullpl{gan} are specific \glspl{ann} architectures that were introduced in 2014 by Ian GoodFellow~\cite{goodfellow2014}. \glspl{gan} are a type of generative models based on game theory where \glspl{ann} are used to mimic a data distribution. Since they were firstly introduced, \glspl{gan} have supposed a large change in the synthesized data generated by \gls{ai}.

Due to their success, the number of \gls{gan} related researches has increased exponentially~\cite{cheng2020generative}. These researches have focused on different aspects of the models, from optimizing their training~\cite{karras2018progressive, gulrajani2017improved} to applying \gls{gan} to new fields such as language generation~\cite{xu-etal-2018-diversity}, image generation~\cite{karras2019stylebased,karras2018progressive}, image-to-image translation~\cite{zhu2017unpaired, isola2018imagetoimage}, image generation in text description~\cite{Zhu_2019_CVPR}, video generation~\cite{Li_Min_Shen_Carlson_Carin_2018}, and other domains~\cite{kim2020learning} achieving state-of-the-art results.

\gls{gan} models are capable of replicating a data distribution and generating synthesized data, applying a certain standard deviation to create new and never seen before data. Due to the particularities of \glspl{gan}, one of the fields were they have supposed a change in the quality of the synthesized data is in computer vision. Although there were previous models~\cite{ackley1985learning, bank2021autoencoders, oord2016pixelrnn}, \glspl{gan} have shown to generate sharper results~\cite{sun2020comparison}.

The main peculiarity of \glspl{gan} lies in their training, where it is based on game theory, where two neural networks compete in a min-max game. Both networks must optimize their corresponding objective functions, generating a situation where two players compete for opposites objectives.

\begin{figure}[h]
    \centering
    \includegraphics[width=.8\textwidth]{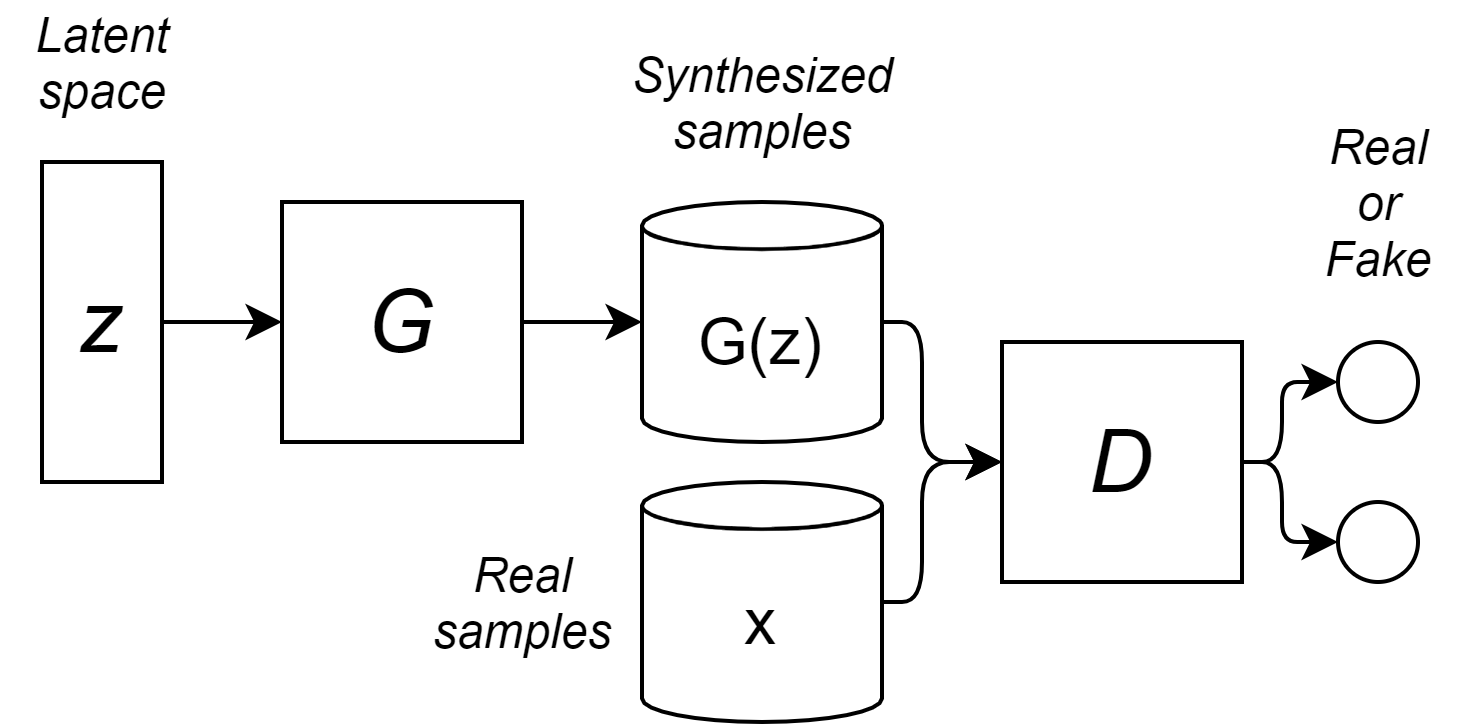}
    \caption{Architecture of a \gls{gan} model.}
    \label{figure:GANArchitecture}
\end{figure}

Fig.~\ref{figure:GANArchitecture} shows how the \gls{gan} architecture is composed. Due to this architecture complexity, \glspl{gan} suffer from instability during their training~\cite{maciej2019stabilizing, thanhtung2019improving, Mao_2017_ICCV}. The instability of training in these models gives rise to problems such as mode collapse, so that researches have been made to tackle this kind of problems~\cite{bhagyashree2020study, bang2018mggan, adiga2018tradeoff, Bau_2019_ICCV, durall2020combating}. As~\cite{tung2020catastrophic} defines, mode collapse happens when the \glspl{gan} model generates the same class outputs with different inputs.

Because of the considerable variety of fields in which \glspl{gan} are applied\cite{AGGARWAL2021generative}, the variety of different \gls{gan} architectures is wide\cite{zhu2017unpaired, isola2018imagetoimage, arjovsky2017wasserstein}. This research focuses on outlining the fields where \glspl{gan} have achieved better results. We will review the different \gls{gan} architectures that exist, how they are structured, and how they are adapted to fulfill the particularities of each problem.

Although we will explain different \gls{gan} architectures, it should be noted that, when new \gls{gan} models are created, they usually combine the different results of previous researches. Most of the models that we will present are usually overlapped to achieve better results.

\gls{gan} surveys are usually focused on \gls{gan} models structure\cite{maciej2019stabilizing, ghosh2020survey} or their application in certain tasks\cite{wang2020generative, alqahtani2019applications}. Because we will focus on novel \gls{gan} architectures, this survey can be identified as the first type. Nevertheless, in the final steps of this survey, we will review how different \gls{gan}~architectures are applied to real world problems.

This survey focuses on contextualizing the recent progress in the \gls{gan} field, reviewing the different variants that have been lately presented and how they address the main problems of training \glspl{gan}. We provide a complete view of the \gls{gan} structure and particularities, then we contextualize the main problems that the networks suffer. We also summarize how \gls{gan} performance is measured, explaining the most used metrics that researchers use. During the different sections we outline how the presented architectures treat the different problems that we have characterized. Finally we propose a classification of \glspl{gan} based on their application, for each class we review the progress that the main variants have followed and we compare their results.

\section{Related Work}

Several other surveys of \glspl{gan} published during the last years \cite{pan2019recent, wang2017generative, ghosh2020survey, sampath2021survey, wu2017survey} have been studied to investigate the recent trends. For example,~\cite{maciej2019stabilizing} focus on the instability issues that \glspl{gan} suffer and show different ways to minimize it. The results suggest that some novel architectures try to control \gls{gan}'s training, while this control can be achieved by focusing on tuning hyper-parameters. It also emphasizes that much of the theoretical work does not fulfill in reality, which causes some \glspl{gan} to convergence when they should not and not converge when they should.

Few surveys have been conducted to explore several approaches to optimize the loss function of \gls{gan}. This research approach tries to enhance the similarity between original and synthesized data distributions by defining an appropriate loss function. Surveys such as \cite{pan2020loss} are focus on analyzing the state-of-the-art \glspl{gan} and further analyzing the performance of a huge variety of networks. In addition, they propose a set of recommendations of which loss function works best for each case of use.

Other works focusing on the applications of \glspl{gan} instead of their composition or loss function. For example,~\cite{gui2020review} focus on how different \gls{gan}'s architectures have been used during the last years for different problems, while~\cite{wang2020generative} shows the different architectures for computer vision and their applications.

Due to the constant evolution of \glspl{gan} during the last few years, these reviews are outdated almost instantaneously. As a result of some relevant and recent researches like~\cite{kim2020learning, zhang2021mffgan, liu2021divco} cannot be found in any recent \gls{gan} review\cite{AGGARWAL2021generative, silva2021review}. We consider that a new and more complete review must be done, covering the researches that previous reviews did not fill in and contributing to a deeper and more thorough analysis of the state-of-the-art of \glspl{gan}.

\section{Structure of this survey}

This survey is structured as follows. Section \ref{section:GAN} is a concise introduction of \gls{gan} composition and principles, we will also summarize the common problems that \glspl{gan} suffer to then review the different solutions proposed to each problem. The different evaluation metrics are also reviewed, we address each metric strengths and weaknesses.

Section \ref{section:GANvariants} reviews the most important \glspl{gan} proposed since their introduction in 2014, paying special attention to the \glspl{gan} proposed in the recent years. This section is divided in two types of \glspl{gan}, the ones focused on improving the architecture of the \gls{gan} and the ones that tries to improve the \gls{gan} performance by changing its loss function behavior. This section also includes a new taxonomy of the reviewed articles and a timeline to have a clear vision of how the research in this field has been developed.

Section \ref{section:GANApplications} summarizes the most important application of \gls{gan} architecture related to computer vision tasks. This section also includes \glspl{gan} applied to other domains different than image generation, paying special attention to the treatment of different types of data, such as molecular composition or medical imaging.

Finally, section \ref{section:Discussion} discuss the actual situation of \gls{gan} with the development of new architectures such as diffusion models or transformers. Here, we describe the potential of these new models in comparison with \gls{gan}.

        \small
        \begin{longtable}{|c|c|c|}
            \hline
            \multicolumn{2}{|c|}{\textbf{Section}} & \textbf{Content} \\
            \hline
            
             \multirow{4}{*}{\ref{section:GAN}} & \textbf{Subsection} & \textbf{\nameref{section:GAN}} \\ \cline{2-3} 
             & \ref{section:DefintionAndStructure} & \nameref{section:DefintionAndStructure} \\ \cline{2-3} 
             & \ref{section:CommonProblemsOfGANs} & \nameref{section:CommonProblemsOfGANs} \\ \cline{2-3} 
             & \ref{section:EvaluationMetrics} & \nameref{section:EvaluationMetrics} \\
             \hline
            
             \multirow{34}{*}{\ref{section:GANvariants}} &  \textbf{Subsection} & \textbf{\nameref{section:GANvariants}} \\  \cline{2-3} 
             
             & \ref{section:ArchitectureOptimisationBasedGANs} & \textbf{\nameref{section:ArchitectureOptimisationBasedGANs}} \\ \cline{2-3}
             
             & \ref{section:dcgan} & DCGAN \\ \cline{2-3}
             & \ref{section:cgan} & CGAN \\ \cline{2-3}
             & \ref{section:acgan} & ACGAN \\ \cline{2-3}
             & \ref{section:infogan} & InfoGAN \\ \cline{2-3}
             & \ref{section:pix2pix} & Pix2Pix \\ \cline{2-3}
             & \ref{section:cyclegan} & CycleGAN \\ \cline{2-3}
             & \ref{section:dualgan} & DualGAN \\ \cline{2-3}
             & \ref{section:discogan} & DiscoGAN \\ \cline{2-3}
             & \ref{section:GANILLA} & GANILLA \\ \cline{2-3}
             & \ref{section:progan} & ProGAN \\ \cline{2-3}
             & \ref{section:dggan} & DGGAN \\ \cline{2-3}
             & \ref{section:stylegan} & StyleGAN \\ \cline{2-3}
             & \ref{section:Alias-FreeGAN} & Alias-Free GAN \\ \cline{2-3}
             & \ref{section:sagan} & SAGAN \\ \cline{2-3}
             & \ref{section:biggan} & BigGAN \\ \cline{2-3}
             & \ref{section:ylgan} & YLGAN \\ \cline{2-3}
             & \ref{section:qugan} & QuGAN \\ \cline{2-3}
             & \ref{section:eqgan} & EQGAN \\ \cline{2-3}
             & \ref{section:cegan} & CEGAN \\ \cline{2-3}
             & \ref{section:ssdgan} & SSD-GAN \\ \cline{2-3}
             & \ref{section:miegan} & MIEGAN \\ \cline{2-3}
             
             & \ref{section:LossFunctionOptimisationBasedGANs} & \textbf{\nameref{section:LossFunctionOptimisationBasedGANs}} \\
             \cline{2-3}
             
             & \ref{section:wgan} & WGAN \\ \cline{2-3}
             & \ref{section:wgangp} & WGAN GP \\ \cline{2-3}
             & \ref{section:ls-gan} & LS-GAN \\ \cline{2-3}
             & \ref{section:lsgan} & lsGAN \\ \cline{2-3}
             & \ref{section:ugan} & UGAN \\ \cline{2-3}
             & \ref{section:RealnessGAN} & Realness GAN \\ \cline{2-3}
             & \ref{section:sngan} & SNGAN \\ \cline{2-3}
             & \ref{section:csgan} & CSGAN \\ \cline{2-3}
             & \ref{section:missgan} & MISS GAN \\ \cline{2-3}
             & \ref{section:SphereGAN} & Sphere GAN \\ \cline{2-3}
             & \ref{section:srgan} & SRGAN \\ \cline{2-3}
             & \ref{section:wsrgan} & WSRGAN \\ \cline{2-3}
             
             & \ref{section:GANTimeline} & \textbf{\nameref{section:GANTimeline}} \\ \hline

             \multirow{12}{*}{\ref{section:GANApplications}} & \textbf{Subsection} & \textbf{\nameref{section:GANApplications}} \\ \cline{2-3} 
             
             & \ref{section:ImageSynthesis} & \nameref{section:ImageSynthesis} \\ \cline{2-3} 
             & \ref{section:Image-to-imageTranslation} & \nameref{section:Image-to-imageTranslation} \\ \cline{2-3} 
             & \ref{section:VideoGeneration} & \nameref{section:VideoGeneration} \\ \cline{2-3} 
             & \ref{section:ImageGenerationFromText} & \nameref{section:ImageGenerationFromText} \\ \cline{2-3}
             & \ref{section:Language generation} & \nameref{section:Language generation} \\ \cline{2-3}
             & \ref{section:DataAugmentation} & \nameref{section:DataAugmentation} \\ \cline{2-3}
             & \ref{section:OtherDomains} & \textbf{\nameref{section:OtherDomains}} \\ \cline{2-3}
             & \ref{section:gamegan} & \nameref{section:gamegan} \\ \cline{2-3}
             & \ref{section:medicalgan} & \nameref{section:medicalgan} \\ \cline{2-3}
             & \ref{section:agriculturegan} & \nameref{section:agriculturegan} \\ \cline{2-3}
             & \ref{section:druggan} & \nameref{section:druggan} \\ \hline
             
             \multicolumn{2}{|c|}{\ref{section:Discussion}} & \textbf{\nameref{section:Discussion}} \\
             
            \hline
        \caption{Summary of the survey}
        \label{tab1}
        \end{longtable}


\section{\acrfullpl{gan}}
\label{section:GAN}

In this section, we will review the basic characteristics of \glspl{gan}, their structure, composition, and common problems. We will especially focus on \gls{gan} problems because most of the \gls{gan} architectures~\cite{metz2017unrolled, suh2021cegan} are created to minimize the training problems.

\subsection{Definition and structure}
\label{section:DefintionAndStructure}

\glspl{gan} are an architecture composed of various neural networks, their objective is to replicate a data distribution in an unsupervised way. To achieve it, they are composed of two neural networks that play a two-player zero-sum game. In this game, the network called the Generator (G) is in charge of creating new data samples replicating, but not copying, the origin data distribution; while the Discriminator (D) tries to distinguish real and generated data.

From a formal point of view, D estimates $p ( y | x )$, that is, the probability of a label $y$ given the sample $x$; while G generates a sample given a latent space $z$, which can be denoted as $G(z)$.

This process consists in both networks competing. While G tries to generate more realistic results, D improves its accuracy detecting which samples are real and which not. In this process, both competitors are synchronized, if G creates a better output, it will be more difficult for D to differentiate them. On the other hand, if D is more precise, it will be more difficult for G to fool D. This process is a minimax game in which D tries to maximize the accuracy and G tries to minimize it. The formulation of the minimax game loss function can be denoted as:

\begin{equation}
\begin{aligned}
    \min_{G}\max_{D} L(D,G) = E_{x\sim p_{r}} log[D(x)] + E_{z\sim p_{z}} log[1 - D(G(x))]
\end{aligned}
\end{equation}

where $ x\sim p_r $ is the distribution of the real data and $ z\sim p_z $ denotes the probability distribution of the latent space of G. $ z\sim p_z $ is commonly a Gaussian or uniform noise that G uses to model new samples of data denoted as $G(z)$. D function is to differentiate between the real distribution $D(x)$ and the synthesized distribution $ D(G(x))$.

According to the equation, the initial publication where the \glspl{gan} where presented\cite{goodfellow2014} proved the existence of a unique solution. This solution is called \gls{ne} and it happens when neither player can improve their loss\cite{nash1951non}.

Several researches have demonstrated that reaching the \gls{ne} might not be possible in practice\cite{farnia2020gans, heusel2017gans} or the unique solution\cite{talavera2021dynamics}.

\subsection{Common problems}
\label{section:CommonProblemsOfGANs}

Due to \gls{gan}'s particularities previously described, there are some aspects in \gls{gan}'s training\cite{salimans2016improved} to which special attention should be given.

In addition of summarizing the different main \glspl{gan} problems, during the section \ref{section:GANvariants} we will connect the different \gls{gan} architectures with the problems that they tackle. It should be noted that the recent proposed architectures tries to minimize the different \gls{gan} issues to optimize their models.

\subsubsection{Mode collapse}

The objective is to generate synthesized data from a latent space, which requires not only quality in the generated data, but generalization and diversity in the different synthesized samples. In other words, \gls{gan} models should be able to recreate new unseen data. Mode collapse occurs when the same class outputs are generated by different inputs from the latent space~\cite{zhang2019towards}.

There are studies~\cite{adiga2018tradeoff} that shows how the quality and diversity of \glspl{gan} are correlated. Many efforts~\cite{demeulemeester2021bures, bang2018mggan, li2021tackling} have been taken to tackle mode collapse, but it is still an open problem.
 
In practice, it is not common for \gls{gan}'s model to generate always the same output with different inputs\cite{goodfellow2017nips}, this issue is known as \textit{complete mode collapse}. This type of error occurs rarely, however, it is a common problem that occurs in a partial form or partial mode collapse, in which a high number of outputs are identical. For example, in image generation, partial mode collapse happens when different outputs contain the same color or texture. It has been proven\cite{goodfellow2017nips} that mode collapse lacks the convergence of the \glspl{gan} even when \gls{ne} is found.

Many of the recently proposed \gls{gan} variants tries to reduce the mode collapse problem. For example there has been proven that \gls{wgan} reduces mode collapse~\cite{bang2018mggan, pei2021dp, li2021tackling}.

\subsubsection{Gradient vanishing}

\gls{gan}'s training must be balanced, both G and D need to be synchronized to learn together progressively~\cite{su2018ganqp, zhang2019towards}. A very accurate D is capable to differentiate between the real and synthesized data, this can be denoted as $ D(x) = 1 $ and $ D(G(z)) = 0$. The loss function in this case approaches to zero, generating gradients close to zero and providing little feedback to the G. On the other hand, a poorly accurate D cannot differentiate between real and synthesized data, providing to G useless information.

\subsubsection{Instability}

Due to the particularity of \glspl{gan}, the combination of two models learning from each other is a complex task. \gls{gan} training is based on a zero-sum game where both networks compete to find its particular solution, playing a minimax game.

This architecture of models is based on cooperation to optimize the global loss function, but the problems that D and G must optimize are opposite. Due to the particularity of the objective function of the networks, there can be times during the training where a small change in one of the networks can lead to a big change in the other, in turn producing further changes. Those intervals in which both networks start to desynchronize their states are very delicate since large changes in the gradients can lead to a network losing its learning~\cite{arjovsky2017wasserstein, Zuo2021WACV}.

It should be noted that instability periods tend to generate more instability, making the problem last longer. Networks can reverse the instability process, but even if it happens, it will cost the training performance.

Many of the last proposed \gls{gan} architectures are focused on stabilize their training~\cite{karras2018progressive, arjovsky2017wasserstein}. By stabilizing the training, it is usually achieved a better performance of the networks, this is why most of the last progress involve a more stable training.

\subsubsection{Stopping problem}

Traditional neural networks have to optimize a loss function decreasing monotonically, in theory, the cost function. Due to the minimax game that \glspl{gan} have to optimize, this does not happen to them~\cite{talavera2021dynamics, liu2017approximation, barnett2018convergence}. In a \gls{gan} training, the loss function does not follow any pattern, so it is not possible to know the state of the networks by their loss function. This causes that, when a training is occurring, it is not possible to know when the models have been fully optimized.

\subsection{Evaluation metrics}
\label{section:EvaluationMetrics}

Due to \gls{gan}'s particularity, there is not an unique metric to measure the quality of the synthesized data~\cite{xu-etal-2018-diversity}. One of the reasons of why there is no consensus among researches is the particularity of each \gls{gan} application. As mentioned in previous sections, \glspl{gan} can be used to replicate any data distribution, but it depends on the particular problem how to measure the differences between the origin and synthesized distributions~\cite{borji2019pros}.

As there is not an unique universal metric to measure the performance of these kinds of models, during the last years there has been developed different metrics. Each metric has its particular strength and it should be noted that, in practice, different metrics are used and compared to measure different aspects and to have a wider view of the \gls{gan} performance\cite{goodfellow2017nips}.

Since there is not an evaluation metric that fulfills all \gls{gan} possible applications, we will review the most widely used metrics:

\subsubsection{\gls{is} and its variants}

\gls{is}~\cite{salimans2016improved} measures the quality and diversity of the generated samples of a \gls{gan}. To do so, it uses a pretrained neural network classifier called \textit{Inception v3}~\cite{szegedy2015rethinking}. The model is pretrained using a dataset of real world images called \textit{Imagenet}~\cite{deng2009imagenet}, it can differentiate between 1.000 of classes of images.

The \gls{is} is calculated by predicting the probabilities of the generated samples. A sample is classified strongly as one specific class means that it has high quality. In other words, it is assumed that low entropy and high quality data are correlated. The \gls{is} value varies between 1 and the number of classes of the classifier.

One of the main problems of the \gls{is} is that it cannot handle mode collapse. In this case, all generated samples by the \gls{gan} will be practically the same, but the \gls{is} would be very high if the images are strongly classified as one class. If this happens, the \gls{is} could be high and the real situation is very bad.

Other particularity of this metric is that it is designed to measure the quality of images since it uses an image classifier.

Based on \gls{is}, there are some modifications to the metric. For example, \gls{ms}~\cite{nowozin2016fgan} is a evaluation metric that takes into account the prior distribution of the labels over the data, \ie it is designed to reflect the quality and diversity of the synthesized data simultaneously.

Other modification of \gls{is} is the \gls{mis}~\cite{gurumurthy2017deligan}. It measures the diversity within the same class category output, trying to mitigate the mode collapse problem.

Some of them, like \gls{fid}~\cite{heusel2017gans} calculate the mean and covariance of the synthesized images and then calculate the distance between the real and generated image distribution. The distance is measured using the \textit{Fréchet distance}, also known as the \textit{Wasserstein-2 distance}. The \gls{fid} is calculated as follows:

\begin{equation}
\begin{aligned}
    \gls{fid} = |\mu - \mu_{w}|^{2} + tr(\Sigma + \Sigma_{w} - 2(\Sigma\Sigma_{w})^{1/2})
\end{aligned}
\end{equation}

where $ w $ denotes the synthesized data of the G.

The \gls{fid} is the most common used metric to measure the quality of generated images\cite{karras2019stylebased,karras2018progressive,karras2021alias,daras2020your}. The use of a common metric for different architectures allows to compare different results using a common metric. In further sections we will go through different results comparing them using \gls{fid}.

One of the strengths of using this metric is that it takes into consideration contamination such as Gaussian noise, Gaussian blur, black rectangles, swirls, among others.

\subsubsection{\gls{msssim}}

is based on the comparison between two image structures, luminance and contrast at different scales\cite{wang2003multiscale}. The \gls{msssim} provides a metric that compares the similarity between the real and the synthesized dataset. One of the strengths of \gls{msssim} is that it correlates closer pixels with strong dependence. In comparison with other metrics such as \gls{mse}, that calculates the absolute error of an image, \gls{msssim} provides a metric based on the geometry and structure of the image.

The \gls{msssim} scale is based on \gls{ssim}, and this metric is calculated as follows:

\begin{equation}
\begin{aligned}
    SSIM(x,y) = [l_{M}(x,y)]^{\alpha_M} \\
    \cdot \prod_{j=1}^{M}[c_{j}(x,y)]^{\beta_j}[S_j(x,y)]^{\gamma_j}
\end{aligned}
\end{equation}
where $ x $ and $ y $ are two windows of image of common size, $ l $ is the luminance of an image, $ c $ the contrast and $ S $ the structure. The value of \gls{ssim} is a decimal between 0 and 1, the value of 1 represents two identical sets of data. Therefore, it is assumed that the higher value of \gls{ssim}, the higher quality of the synthesized images.

\gls{msssim} is calculated using the average pairwise of \gls{ssim} with N batches. This metric is commonly used with \gls{is} or its variations~\cite{kurach2019the} to provide a wider view of the generated data quality.

\subsubsection{\gls{c2st}}

To measure the quality of the generated distribution, a binary classifier can be used~\cite{lehmann2006testing}. The classifiers divide the samples into synthesized and real ones, judging whether different samples belong to the same data distribution.

It should be noted that this method is not constrained to image evaluation, since a classifier can be used to classify any given data distribution, it can be adapted to any type of input data.

\gls{1nn}~\cite{lopezpaz2018revisiting} is a type of binary classifier used to evaluate \gls{gan} performance. \gls{1nn} is a variant of \gls{c2st} that does not require hyper-parameter tuning. \gls{c2st} using \gls{1nn} is known as \gls{c2st}-\gls{1nn}.

Neural networks can be used as a \gls{c2st}, as mentioned in previous sections, D is indeed a classifier of real and generated data. As is proposed in \cite{lopezpaz2018revisiting}, a \gls{c2st} can be applied to \glspl{gan} by using the same composition of the discriminator, as is said in the paper ``\textit{training a fresh discriminator on a fresh set of data}''. \gls{c2stnn}.

Using \gls{c2st}, we can measure the distance between the synthesized and real data distributions. This provides a useful, human-interpretable metric of \gls{gan} performance. \gls{c2st} has been applied to different \glspl{gan} architectures such as DCGAN or CGAN, using \gls{c2stnn} and \gls{c2st}-\gls{1nn}~\cite{lopezpaz2018revisiting}.

\subsubsection{Perceptual path length}

Using the well-known neural network classifier VGG16~\cite{simonyan2015verydeep} the perceptual path length was designed~\cite{karras2019stylebased} to measure the entanglement of images. The embeddings of consecutive images are calculated using VGG16, interpolating random latent space inputs, then it is calculated how the synthesized images changes.

Drastic change means that, for a minimum change in the latent space there are multiple features that are changing, that means that those features are entangled under the same representation. This metric measures how well the \gls{gan} is learning the different features of the input images, measuring the entanglement of the generated images.

\subsubsection{\gls{mmd}}

is used to measure the distance between two distributions~\cite{bounliphone2016test}. A lower score for \gls{mmd} means that the distributions that are being compared are closer, and that means that the synthesized data is similar to the original.

Given distributions $ \mathbb{P} $ and $ \mathbb{Q} $ and a kernel $ k $. As it is defined in \cite{li2017mmd}, \gls{mmd} can be denoted as:

\begin{equation}
\begin{aligned}
    M_{k}(\mathbb{P}, \mathbb{Q}) = ||\mu_{\mathbb{P}} - \mu_{\mathbb{Q}}||^{2}_{\mathcal{H}} = E_{\mathbb{P}}[k(x, x^{'})] \\
    - 2E_{\mathbb{P,Q}}[k(x, y)] + E_{\mathbb{Q}}[k(y, y^{'})]
\end{aligned}
\end{equation}

It should be noted that this method can be used with any type of data.

\subsubsection{\gls{hr}}

Human classification can be useful in some cases. Either to complement other evaluation metrics, either because there is not other metric that fulfills the particular problem, human evaluation of the generated data can be done.

Due to the particularity of this method, it can only be used when the synthesized data is comprehensive for a human.

For example, in~\cite{zhu2017unpaired, isola2018imagetoimage} human classifications were applied via \gls{amt} to evaluate the realism of the outputs of the \gls{gan}. In this case, participants had to differentiate between the generated and real images. The more images that fool humans perception, the better.

This method can provide an approximation of how \glspl{gan} creation would be perceived by humans.

\section{\gls{gan} variants}
\label{section:GANvariants}

Since the first \gls{gan} was developed~\cite{goodfellow2014} there has been published many different variations of it~\cite{karras2019stylebased,isola2018imagetoimage,karras2018progressive,arjovsky2017wasserstein}. To have a broad vision about recent \gls{gan} researches, we will review the recent progress in this field.

This section is divided into \gls{gan} models according to their main features. That said, we will divide the different \gls{gan}'s variations in architecture modification based and loss function modification based.

\subsection{Architecture optimization}
\label{section:ArchitectureOptimisationBasedGANs}

Some recent researches \cite{karras2019stylebased,karras2018progressive,zhu2017unpaired} are focused on the architecture of the \gls{gan} is designed. Some of them\cite{karras2018progressive} suggest a change in \glspl{gan} training, others\cite{karras2019stylebased} add changes to the structure of the G or D models.

Despite this, we will review traditional \gls{gan}'s architecture, we will focus on models that are relevant for \gls{gan} recent development. It should be noted that the collection of architectures that we will review should not be considered individually.

\gls{gan} model evolution is supported by constant optimization. Therefore, to have a complete vision of \gls{gan} evolution, we will go through the different models that have been relevant in the last years.

\subsubsection{\gls{dcgan}}
\label{section:dcgan}
One year after, the first \gls{gan} was proposed in 2014~\cite{goodfellow2014}, the \gls{dcgan} was introduced\cite{radford2016unsupervised} suggesting some changes to the original architecture. The main objective of the \gls{dcgan} is to use convolutional layers instead of the firstly proposed fully connected layers.

The main change to the fully connected \glspl{gan} is the substitution of the dense layers by convolutional layers. Convolutional layers have been used during the last decade for computer vision tasks. By applying different filters to the images, the convolutional layers are able to extract the main features of the matrix of pixels keeping the correlation between adjacent pixels.

Convolutional layers are used not only used for image processing, but there are recent projects~\cite{jumper2020high} that use matrices of data to take advantage of using convolutional layers.

In addition to the convolutional layers, other changes were suggested to stabilize the \gls{gan}'s training. Replacing the pooling layers by strided convolution has shown better performance~\cite{springenberg2015striving, ayachi2020strided}. Therefore, it is proposed to use strided convolutions in both G and D.

The use of batch normalization layers in both G and D is proposed, this has been shown to reduce the noise and improve the diversity of the generated samples~\cite{yanchun2018improved, shuang2019l1norm}.

To activate the convolutional layers, it is proposed to use a \gls{relu} activation for the hidden layer of G, hyperbolic tangent (tanh) for the output layer of G and leaky rectified linear unit \gls{leakyrelu} for D.

In addition to the mentioned changes in the architecture of the \glspl{gan}, the \gls{dcgan} paper also presents a technique to visualize the filters learned by the models. This helps the comprehension of \glspl{gan} learning methods, confirming previous works related to biology~\cite{hubel1959receptive}.

This architecture supposes a change in how \glspl{gan} are designed and trained. The innovations that were proposed in the paper are applied in most of the following \gls{gan} models.

\subsubsection{\gls{cgan}}
\label{section:cgan}
Proposed in 2014~\cite{mirza2014conditional}, the \gls{cgan} architecture adds a latent class label $ c $ along with the latent space. The new label is used to split the processed data into different classes, thus the synthesized data is generated according to the class of the input label. There are some problems that require the generated data to be classified into different classes~\cite{loey2020deep, ma2018speckle, li2020sar}.

Despite being a simple technique, it has proven to prevent mode collapse. However, the training of a \gls{cgan} requires a labeled dataset complicating its application to some problems.

\gls{cgan} architecture has influenced \glspl{gan} model since its proposition, there has been developed many variations~\cite{isola2018imagetoimage, chen2016infogan, odena2017conditional}.

\subsubsection{\gls{acgan}}
\label{section:acgan}
\gls{acgan}~\cite{odena2017conditional} modifies the \gls{cgan} structure. The D of the \gls{acgan} does not receive the class label $ c $ as an input, instead D is used to classify the probability of the image class. To train the model, the loss function must be modified, dividing the objective function in two parts, one for the correct source of data and the other for the class label. \gls{acgan}~loss function can be denoted as:

\begin{equation}
\begin{aligned}
    L_{s} = E[log P(S = real | X_{real})] \\
    + E[log P(S = fake | X_{fake})]
\end{aligned}
\end{equation}

\begin{equation}
\begin{aligned}
    L_{c} = E[log P(C = c | X_{real})] \\
    + E[log P(C = c | X_{fake})]
\end{aligned}
\end{equation}

where $ L_{s} $ is the log-likelihood of the correct data distribution and $ L_{c} $ is the log-likelihood of the correct class label.

\subsubsection{\gls{infogan}}
\label{section:infogan}
One of the mentioned deficiencies of conditional \glspl{gan} was the requirement of a labeled dataset. \gls{infogan}~\cite{chen2016infogan} provides an architecture to train conditional \glspl{gan} with an unsupervised method. To do so, the latent class label $ c $ is substituted by a \textit{latent code} vector.

The latent space and the latent code are maximized by using the Mutual Information~\cite{shannon1948mathematical}. The mutual information term is not easy to calculate because it requires the posterior $ P(c|x) $. To optimize the training performance, an auxiliary distribution $ Q(c|x) $ is defined. Said so, the loss function of the \gls{infogan} is defined as follows:

\begin{equation}
\begin{aligned}
    \min_{G,Q}\max_{D} V_{InfoGAN}(D,G,Q) = V(D,G) \\
    - \lambda L_{I}(G,Q)
\end{aligned}
\end{equation}

where $ \lambda $ is a hyperparameter that is in charge of the latent code control. As it is proposed in the original paper~\cite{chen2016infogan} a $ \lambda $ equal to 1 is used when the latent code is discrete, for continuous latent codes a smaller $ \lambda $ should be used. The reason for that is to control the differential entropy.

\subsubsection{\gls{pix2pix}}
\label{section:pix2pix}
The main objective of the \gls{pix2pix}~\cite{isola2018imagetoimage} architecture is to do an image-to-image translation. That is, given an image from a domain $ A $, transform this image to other domain $ B $. For example, given a map of a street, transform the map to an aerial photo of the street on the map.

The \gls{pix2pix} architecture is based on an autoencoder, but skips some connections. This architecture is known as U-Net, and it is based on the idea of retrieving information at early stages of the network. The same approaches of skipping connections have been used before~\cite{he2016deep, szegedy2015going, zhou2016learning, johnson2016perceptual} showing great results and improving the network performance.

In addition to the new architecture, a new loss function is proposed that is denoted as:

\begin{equation}
\begin{aligned}
    L_{GAN}(G,D) = E_{y}[]log D(y)] \\
    + E_{x,z} [log(1-D(G(x,z))]
\end{aligned}
\end{equation}

As a follow-up of \gls{pix2pix}, \gls{pix2pix}HD was proposed~\cite{liu2017high} improving the quality of the generated images. Many later works have used \gls{pix2pix}~\cite{qu2019enhanced, mori2020feasibility, pan2021stochastic, drob2021rf} converting it to one of the most popular architectures of the last decade.

The immediate application of these algorithms to images has had a great impact on society, radically increasing its popularity thanks to the applications developed.

\subsubsection{\gls{cyclegan}}
\label{section:cyclegan}
Cyclic consistency is the idea that, given a data $ x $ from a domain $ A $, if the data is translated to a domain $ B $ and translated again to the $ A $ domain it should be recovered the data $ x $. In other words, if a sample is translated to a domain and recovered from that domain, it should not change. This process, where a data sample is transformed and recovered, is known as cycle consistency, and it has been widely used during the last decades~\cite{sundaram2010dense, kalal2010forward}.

This idea is the main base of \gls{cyclegan}~\cite{zhu2017unpaired}. The main strength of the application of cycles is that paired data is not a requirement. \gls{gan} architecture adds a new mapping denoted as F, its function is to do the inverse mapping to retrieve the original data. In other words, the function of F is $ F(G(x)) = x $. To train the architecture, a new cycle consistency loss is proposed to train the so-called \textit{forward} and \textit{backward cycle consistency}. The cycle consistency loss is denoted as follows:

\begin{equation}
\begin{aligned}
    L_{cycle}(G,F) = E_{x~p_{data}(x)}[||F(G(x)) - x||_{1}] \\
    + E_{y~p_{data}(y)}[||G(F(y)) - y||_{1}]
\end{aligned}
\end{equation}

Despite \gls{cyclegan} was first proposed for image-to-image translation, it can be used for any data translation.

\subsubsection{\gls{dualgan}}
\label{section:dualgan}
The architecture of DualGAN~\cite{yi2017dualgan} is very similar to \gls{cyclegan}. As it was with the \gls{cyclegan}, the \gls{dualgan} does not require paired data to train its models. To learn the translation from one data domain to another, \gls{dualgan} has two pairs of identical G and D, each pair is responsible for their respective translation.

To stabilize the training and prevent mode collapse, the loss format of \gls{wgan}~\cite{arjovsky2017wasserstein} is used. This marks the architecture of the network and the construction of the objective function.

In order to train each pair of G and D a \textit{reconstruction error term} is defined. The reconstruction error objective is the same that it was in \gls{cyclegan}, calculating the distance between the original sample of data and its corresponding recovered sample. The reconstruction error is defined as:

\begin{equation}
\begin{aligned}
    l^{g}(u,v) = \lambda_{U}||u - G_{B}(G_{A}(u, z), z')|| \\
    + \lambda_{V}||v - G_{A}(G_{B}(v, z'), z)|| \\
    - D_{B}(G_{B}(v, z')) - D_{A}(G_{A}(u, z))
\end{aligned}
\end{equation}

while $ U $ and $ V $ are both domains, $ \lambda_{U} $ and $ \lambda_{V} $ are two constant parameters and $ z $ and $ z' $ are both random noises. $ \lambda_{U} $ and $ \lambda_{V} $ are normally set a value within [100.0, 1, 000.0], when the domain U contains real images (e.g. a human face photo) and V does not (e.g. a sketch of human face), it is more optimal to use a smaller value of $ \lambda_{U} $ than $ \lambda_{V} $.

\gls{dualgan} has been widely used and modified~\cite{ye2020data, prokopenko2019synthetic, liang2021improved}. For example, in \cite{veillon2021towards} a \gls{dualgan} architecture was used to transform an input speech emotion. In this application, given the \gls{f0} of a certain emotion, the trained network is capable of changing the emotion of the sound. To do so, \gls{f0} is encoded using wavelet kernel learning~\cite{yger2011wavelet} using the same methodology as~\cite{luo2019emotional}.

\subsubsection{\gls{discogan}}
\label{section:discogan}
\gls{discogan}~\cite{kim2017learning} is an architecture that follows the same structure as \gls{dualgan} and \gls{cyclegan}. The particularity that \gls{discogan} has is the usage of an autoencoder for the G. For D, it uses a classifier based on the encoder of the G.

Autoencoders have been used to other reconstruction problems~\cite{chaitanya2017interactive, luchnikov2019variational, mehta2017rodeo}, so applying of this architecture to domain-to-domain translation problems can benefit from their particularities. Autoencoders are based on the idea of reducing the dimensionality of the input data, then they reconstruct the same information. By doing the dimensional reduction, the network is capable of maintaining the essential features of the input data. In the case of domain-to-domain translation, by using autoencoders, the architecture is capable of maintaining the main features of a sample and translating this core information to other specific domains.

The results presented in the original work show how \glspl{gan} can learn high-level relationships between two complete different domains. In the experiments carried out in the research, it was demonstrated how the networks discovered relationships such as orientation. E.g., pairing images of chairs and car with the same orientation.

\subsubsection{GANILLA}
\label{section:GANILLA}
The GANILLA~\cite{hicsonmez2020ganilla} architecture modifies the structure of the G of the \gls{gan} for image style transfer. The main objective of the variant is to maintain both the content and the style of an image, previous methods usually lack one of this aspects in favor of the other. The main idea of the GANILLA is to do the style transfer of an image balancing style and content.

The architecture of GANILLA uses low-level features to maintain the content of the image at the same time as the style transfer is done. The G model is based on two stages, one for downsampling the input image and the other for upsampling the information of the first stage. This architecture ensures that the style transfer maintains the input features of the image but, in addition, some layers concatenate features of previous layers such as edges, shapes or morphological features. With these two methods, the architecture controls both content and style.

\begin{figure}[h]
    \centering
    \includegraphics[width=\textwidth]{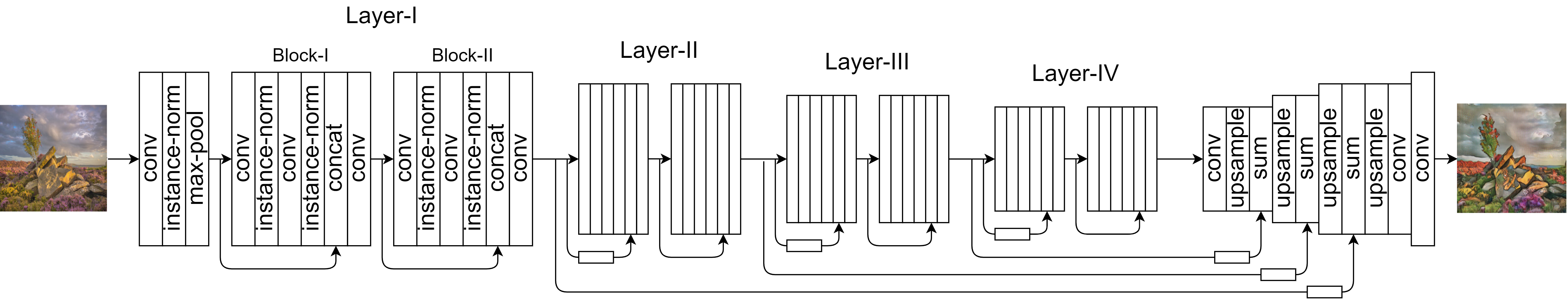}
    \caption{Structure of the proposed architecture of the GANILLA. Figure based on Reference \cite{hicsonmez2020ganilla}.}
\label{figure:GANILLA}
\end{figure}

The downsampling stage is based on ResNet-18~\cite{he2016deep} but with skipped connections. This skipped connections then feed the upsampling module. The architecture of the GANILLA can be observed in the Fig.\ref{figure:GANILLA}

For training the models, the cyclic consistency method of the \gls{cyclegan}~\cite{zhu2017unpaired} is used. This way, two pairs of G and D are used to map both domains.

The results of the GANILLA show the good performance, in specific for children's book illustration dataset. Due to the particularities of the images of children's books, being highly contrasted images with abstract objects, previous architectures had difficulty to do the style transfer. However, with the usage of low level features of the GANILLA, it is achieved an improvement of the overall performance.

\subsubsection{\gls{progan}}
\label{section:progan}
Training a complex model can lead to strong instability. To tackle the instability of \glspl{gan} models, \gls{progan}~\cite{karras2018progressive} proposes a training methodology based on a growing architecture. The idea of a progressive neural network was previously proposed~\cite{rusu2016progressive}.

\begin{figure}[h]
    \centering
    \includegraphics[width=\textwidth]{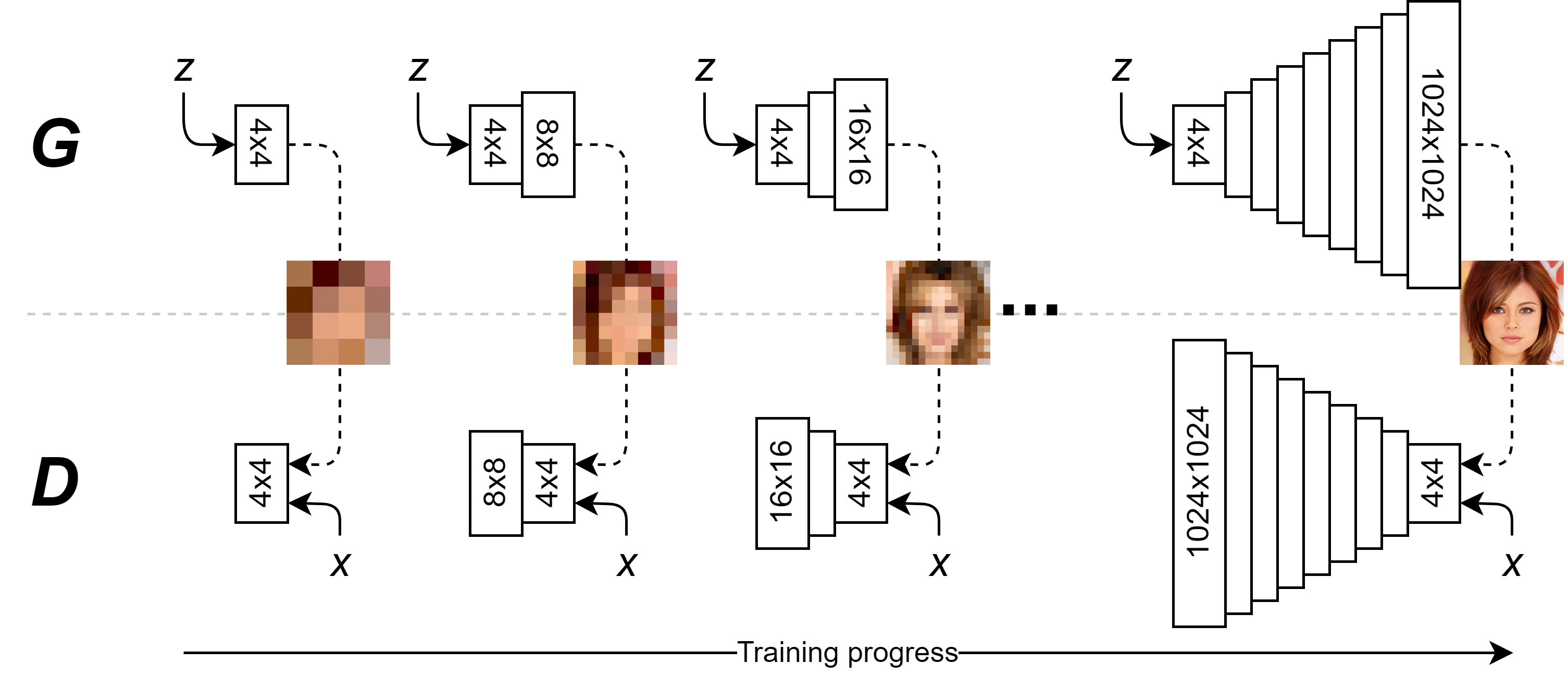}
    \caption{Training schedule of ProGAN. Figure based on Reference \cite{karras2018progressive}.}
    \label{figure:ProGANTraining}
\end{figure}

The main idea behind progressive networks is the concatenation of different training phases. In each phase, a model is trained and, as the trainings are developed, the model number of layers increases. This way the created model scales up gradually stabilizing the training. The strength of this architecture is that, due to the simplicity of the first model, the networks are capable to learn properly the simplest form of the problem and then use the learned characteristics to scale up little by little the complexity of the problem. With each new phase, it is important to emphasize that the weights of the networks remain trainable, letting them to adapt to the new phases. A scheme of the progressive training of \gls{progan} can be seen in Fig.\ref{figure:ProGANTraining}.

Due to the explained training methodology, \gls{progan} is capable to stabilize the training of \glspl{gan}, which is one of the most important \gls{gan} problems. In addition, \gls{progan}'s training methodology speeds up the training phase and produces images of state-of-the-art quality, e.g. achieving an inception score of 8.8 in the unsupervised CIFAR-10~\cite{krizhevsky2009learning} dataset.

The \gls{progan} described in the original paper used the \gls{wgangp}~\cite{gulrajani2017improved} loss format, despite that \gls{progan} architecture can be applied to any loss function. \gls{progan} training methodology has been implemented in many recent researches~\cite{yang2021proegan, bhagat2019data}.

\subsubsection{\gls{dggan}}
\label{section:dggan}
\gls{dggan}~\cite{liu2021dynamically} proposes a new training methodology based on \gls{progan}. The architecture of the networks of \gls{dggan}~not only grow periodically, they rather grow dynamically adapting their architecture and parameters during the training.

The \gls{dggan} questions some aspects of \glspl{gan} such as the symmetry between G and D or layer choice. The new methodology can automatically search the optimal parameters, respecting \gls{progan} growing strategy was previously defined.

The \gls{dggan} starts with a base D and G, the training alternates between training steps and the growing of the network. To grow the network, a set of $ child $ architectures are created. Each child has the same architecture as the parent, but each child proposes a different growing change to the network. During the training children architectures are trained, initializing the weights of the inherited parent layers with their respective parent weights.

\begin{figure}[h]
    \centering
    \includegraphics[width=\textwidth]{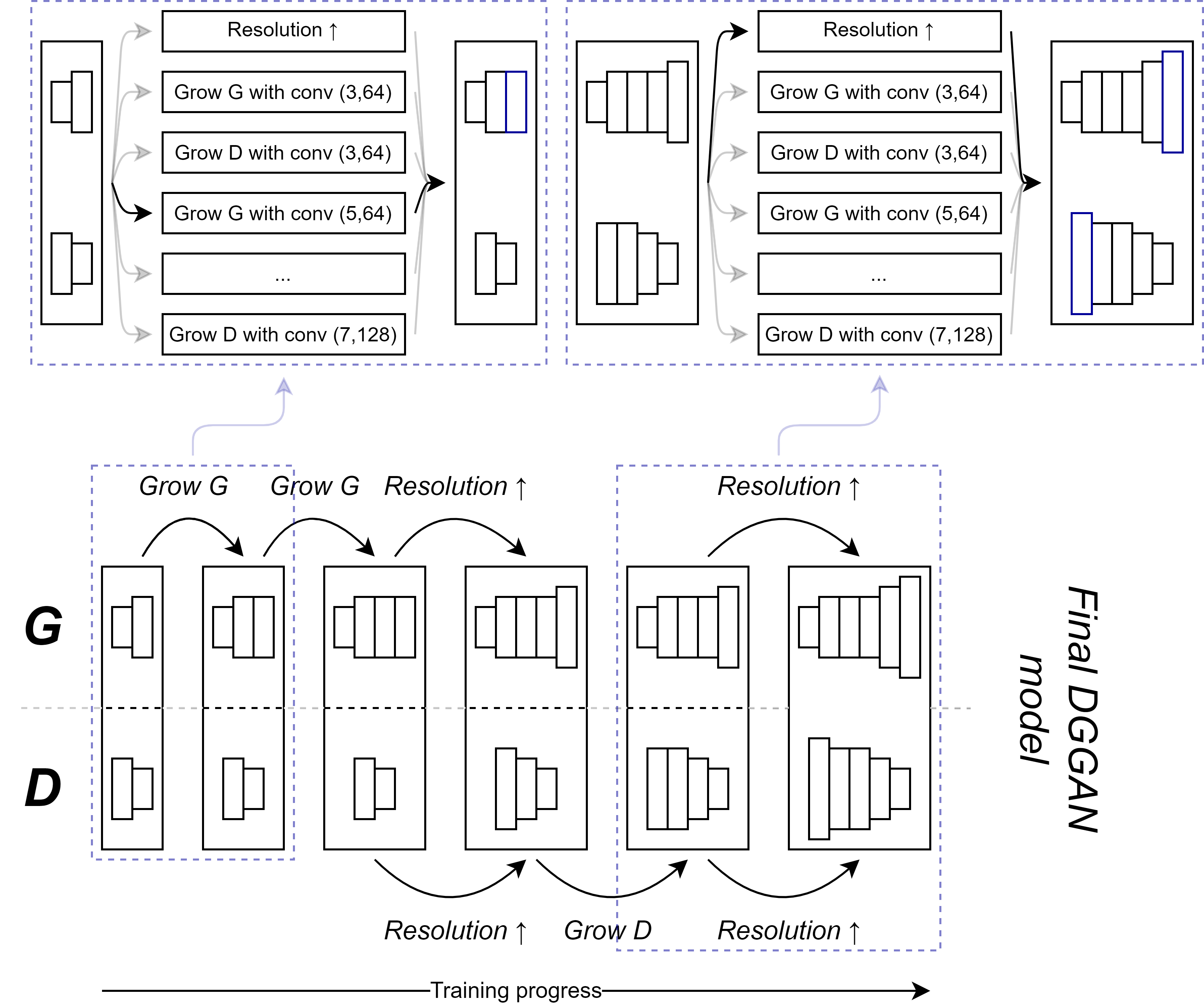}
    \caption{Training methodology of \gls{dggan}. Figure based on Reference~\cite{liu2021dynamically}.}
    \label{figure:DGGANTraining}
\end{figure}

In the proposed dynamic growing algorithm, each step chooses among different growing possibilities: grow G with a certain convolution layer, grow D with a certain convolution layer, or grow both G and D to a higher resolution. A scheme of the training methodology can be seen in Fig.\ref{figure:DGGANTraining}.

If all children were preserved in each step, it will produce an exponential growing that would lead to large inefficiency. To avoid that, before the children generation, a prune is made. Known as greedy prune, the prune is done by keeping the top $ K $ children of each generation. Then each child becomes a parent and generates a new batch of children. The process repeats until the network grows to the desired size.

In the original research, the child search was made combining different kernel sizes and number of filters, each parameter is known as an action, and the number of total actions is denoted as $ T $. It can be easily noted that different hyperparameters can be searched by using this algorithm. To avoid a large increment of the number of children, the algorithm proposes a probability $ p $ of a child to test a new parameter. A higher $ K $, $ T $ and $ p $ means a wider search, contributing to a better exploration of the candidates but a slower training.

It should be noted that the search algorithm lacks the efficiency of the architecture by having to do multiples training simultaneously. It also lacks the ability of growing, due to the quick growing of the number of networks.

\subsubsection{\gls{stylegan}}
\label{section:stylegan}
\gls{stylegan}~\cite{karras2019stylebased} is based on the idea that, improving the processing of the latent space, the quality of the generated data will improve. Due to the particularities of the latent space, there are many interpolations on the variables~\cite{sainburg2018generative, laine2018feature} that produces entanglement in the learned characteristics of the G. The architecture of the \gls{stylegan} is based on previous style transfer researches~\cite{huang2017arbitrary}.

With the architecture of \gls{stylegan}, G is capable to learn different \textit{styles} of the input data disentangling high-level characteristics. This produces an improvement on the quality of the generated data and helps in the interpretation of the latent space, previously poorly understood. Controlling the latent space leads to better interpolation properties, enabling interpolation operations in different scales, e.g., interpolation of poses, hair or freckles in human face images.

In the \gls{stylegan} architecture, the input of G is mapped to an intermediate latent space called $ W $, then is used in each convolution layer via an \gls{adain}. In addition to the latent space, \textit{gaussian noise} is added to the output of each convolution layer.

The \gls{stylegan} architecture uses the training methodology used in \gls{progan}, supporting the previously mentioned idea that each research should not be considered as an isolated result. The paradigm of investigation is supported by the continuous mixing of new techniques.

Said so, the \gls{stylegan} improves the quality of the generated images of the \gls{progan}, achieving a \gls{fid} score of 5.06 in CelebA-HQ dataset and 4.40 in FFHQ dataset.

\subsubsection{Alias-Free GAN}
\label{section:Alias-FreeGAN}
During the last years, multiples architectures have been improving the quality of the synthesized images. The previously mentioned \gls{stylegan} achieved one of the best results in image generation, producing images of human faces with a quality never seen before. Besides its good results, some problems remain opened.

One of the most visible problems that generated images of \gls{stylegan} had was the known as \textit{texture sticking}. It happens when a certain image feature depends on absolute coordinates instead depending on other feature localization. E.g. the texture of the beard of a human face seems stuck when interpolating different images. The texture sticking problem is noticeable especially when interpolating images, e.g. changing the posture of a human face image.

Alias-Free GAN~\cite{karras2021alias} focus on solving the texture sticking problem of the \gls{stylegan}. The main idea is to suppress the alias in the generated images, this way the finer details will be attached to the underlying surface of the image. To achieve this, each layer of G is designed to be equivariant by applying rotations and translations to the continuous input.

To achieve an equivariant G, many changes have been made. A 10-pixel margin is used for the internal representations, due to the assumption of infinite spatial extension for the feature maps. The Leaky \gls{relu} layers are wrapped between an upsampling and a downsampling, this is implemented with a CUDA kernel for optimization. The cutoff frequency of the \gls{stylegan} is cut off to ensure the alias frequencies are in the stopband. In addition, the learned input constant of \gls{stylegan} is substituted by Fourier features~\cite{tancik2020fourier, xu2021positional}. Finally, the rotation equivariant version of the network is obtained by reducing the kernel size of $3 \times 3$ convolutions to $1 \times 1$ and changing the sinc-based downsampling to a radially symmetric jinc-based one.

\subsubsection{\gls{sagan}}
\label{section:sagan}
\gls{sagan}~\cite{zhang2019self} architecture covers the problem of local spatial information of images. I.e. images that have different components correlated in different positions of the image can be difficult to cover because the receptive field of the network is not big enough. In SAGAN, the generation of different features is made considering cues from all images. In addition, \gls{sagan} D is capable of evaluating the consistency of features along the image.

\begin{figure}[h]
    \centering
    \includegraphics[width=\columnwidth]{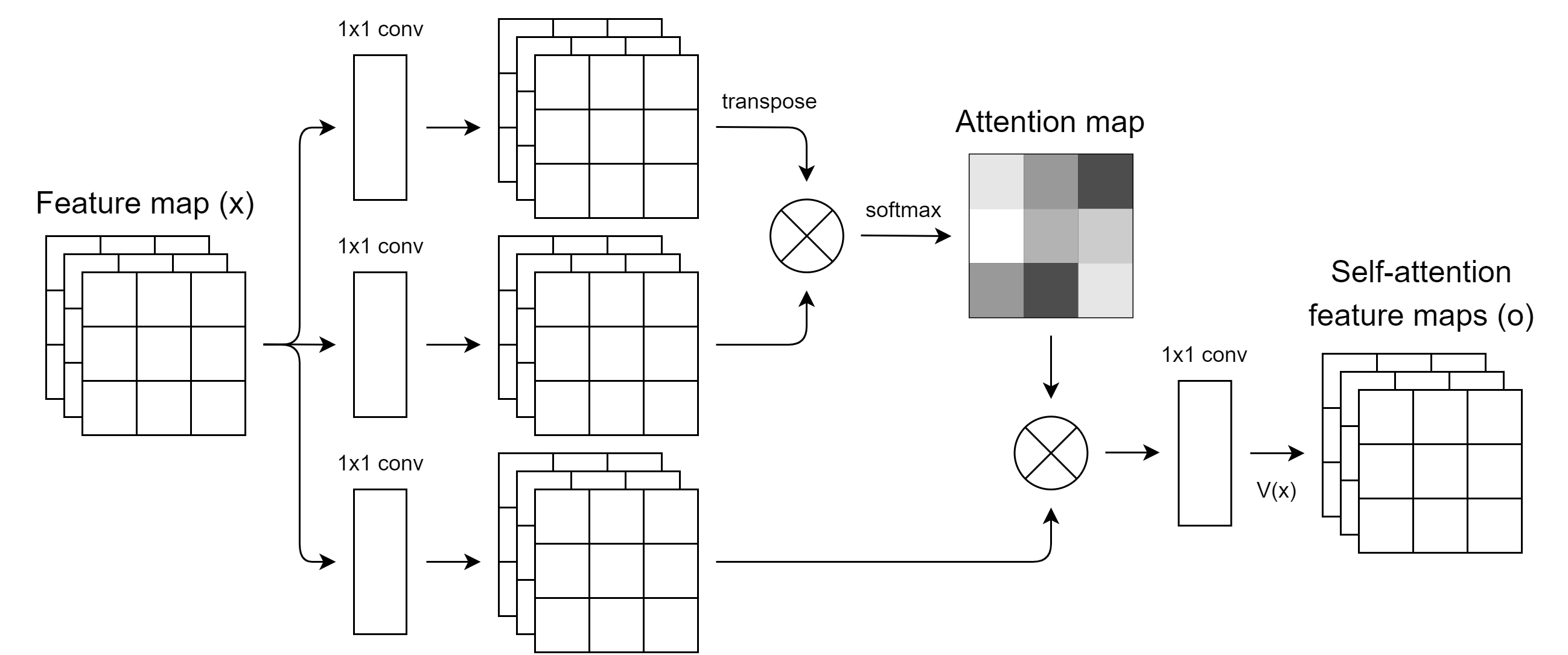}
    \caption{Self attention layer of \gls{sagan}. Figure based on Reference \cite{zhang2019self}.}
    \label{figure:SelfAttentionLayer}
\end{figure}

SAGAN uses self attention layers~\cite{vaswani2017attention}, these layers are capable to capture structural and geometric features of multiclass datasets. The feature maps of each convolution are split into a $1 \times 1$ convolution in \textit{query}, \textit{key} and \textit{value}, then they are multiplied to construct the output of the layer. This way the network can learn long-range dependencies. The structure of the self-attention layer can be seen in Fig.\ref{figure:SelfAttentionLayer}.

\subsubsection{BigGAN}
\label{section:biggan}
The BiGAN architecture~\cite{brock2018large} focuses on generating high resolution images from diverse datasets. Previous models results were able of synthesize new samples of low dimensionality, they had problems when scaling their results to bigger samples. The results achieved by the BigGAN, in terms of \gls{fid} and \gls{is} outperform previous models.

The researches of the BigGAN claim that \glspl{gan} have better performance when they use higher dimensional data. The architecture of the BigGAN is based on the \gls{sagan}~\cite{zhang2019self} architecture. The authors show that, by enlarging the number of channels of the images used by a factor of 50\%, the \gls{is} improve by a factor of 21\%.

One innovation proposed in this article is the so-called "Truncation Trick". Previous \gls{gan} models used a normal or uniform distribution to generate the latent space of the G network. The authors claim that by using a truncated normal distribution the results, in terms of \gls{fid} and \gls{is}, were better. This truncation trick reduce the variety of values of the latent space by truncating them towards zero. The main drawback produced by this is that the variability of the generated samples is reduced. It exists a relationship between the variety and fidelity of the generated samples using this truncation. The more truncation applied to the latent space, the less variety of images were produced.

Other aspect that is scaled up in this work is the batch size of the \gls{gan} training, increasing it by a factor of 8. The authors show that by using larger batches the gradients of each iteration are better, reaching a better performance in less steps. This is caused because the composition of each batch is more diverse, being able of covering more modes of the data.

\subsubsection{\gls{ylgan}}
\label{section:ylgan}
\gls{ylgan}~\cite{daras2020your} proposes a new attention layer that substitutes the \gls{sagan} dense attention layer~\cite{zhang2019self}. This new layer preserves two-dimensional image locality and contributes the flow of information through the different layers. To preserve the two-dimensional locality and quantify how information flows through the model, the framework of \gls{ifd}~\cite{dimakis2010network} is used.

The modification of the self attention layer of \gls{sagan} introduces sparse attention layers. This new method reduces the quadratic complexity of the attention layer by splitting the attention into multiple subsets of data. The main problem of the sparse attention layer is that, besides its computational optimization, it lacks the information flow of the network. To tackle this information flow graphs are introduced, these graphs will be used to support Full Information through the layers of the network.

The results show how applying the new layer improves the quality of the images compared to the \gls{sagan} generated images. The architecture of the \gls{sagan}, modifying the dense attention layer and preserving the rest parameters is called YLG-SAGAN. YLG-SAGAN not only improves the \gls{fid} of \gls{sagan}, reducing it score from 14.53 to 8.95, furthermore it optimizes the training time to around a 40\%.

\subsubsection{\gls{qugan}}
\label{section:qugan}
During the last decade, quantum computing has become a hot topic in computer science. Since it was proposed in 1980~\cite{benioff1980computer} it has always been restricted to a few laboratories around the world. Thanks to the progress made recently~\cite{macquarrie2020emerging}, it has made possible to test the first algorithms, prototypes and ideas\cite{cao2019quantum}.

Thanks to quantum computing particularities, problems previously defined can be solved, or are optimized, reducing their computation time. Using quantum superposition, the multiples solutions can be evaluated simultaneously, then by using quantum interference and entanglement the correct answer can be defined.

\gls{qugan}~\cite{stein2020qugan} proposes a \gls{gan} architecture powered by quantum computing. By using quantum computing, \glspl{gan} are hugely optimized, reducing a 98.5\% of its parameter set compared to traditional \glspl{gan}.

\gls{qugan}~architectures use qubits to create the quantum layers of G and D, known as QuG and QuD. The data that the networks use is transformed into quantum states.

\subsubsection{\gls{eqgan}}
\label{section:eqgan}
\gls{eqgan}~\cite{niu2021entangling} proposes a variation of the previously proposed quantum \glspl{gan}. Benefiting from the entangling properties of quantum circuits, \glspl{eqgan} guarantees the convergence to a \gls{ne}.

The main particularity of \gls{eqgan} is that it performs quantum operations on both synthesized and real data. This approach produces fewer errors than swapping the data between quantum and classical.

To apply \gls{eqgan} to real problems, a \gls{qram} is used. By using the \glspl{qram}, the \gls{eqgan} is capable to improve the performance of the D.

\subsubsection{\gls{cegan}}
\label{section:cegan}
Data imbalance is a common problem when using real world datasets. Dataset often contains a majority of samples of a certain data class. In the case of \glspl{gan} using unbalanced datasets, the imbalance problem results in poor quality of the synthesized data of the class with less samples.

\gls{cegan}~\cite{suh2021cegan} tries to solve the data imbalance problem in \gls{gan}. The objective is to enhance the quality of the synthesized data and to improve the accuracy of the predictions.

The \gls{cegan} architecture consists of 3 different networks, G, D and a new network known as the classifier (C). The training of the \gls{cegan} divides in two steps. In the first step, the architecture is normally trained, using D to differentiate between fake and real samples, C is used to classify the class label of the input sample. Then, in the second step, an augmented training dataset is formed via generating new samples from G, and this new dataset is used to train the C.

The methodology presented in \gls{cegan} substitutes previous techniques to deal with data imbalance. Unlike other methods such as undersampling~\cite{ng2014diversified} or oversampling~\cite{ramentol2012smote} \gls{cegan} does not modify the original dataset. This way, some problems of the traditional methods are avoided, e.g. shortening the original dataset by undersampling or redundant information by oversampling with geometric transformations. 

\subsubsection{\gls{ssdgan}}
\label{section:ssdgan}
The \gls{ssdgan}~\cite{chen2020ssd} tackles the problem of high frequency samples in \glspl{gan}. The described problem causes high spectrum discrepancies between the real and the synthesized samples. The \gls{ssdgan} proposes to alleviate this discrepancy to enhance the quality of the synthesized data.

The idea behind the architecture is to reduce the gap of spectrum discrepancy, combining the spectral realness and the spatial realness of each sample, to do so a new D is defined. The new proposed D is known as $ D^{ss} $ and it combines D and a classifier C. D is in charge of measuring the spatial realness of an image, this is the same approach of the D of the traditional \gls{gan}~\cite{goodfellow2014}. The new proposed C is in charge of the known as \textit{spectral classification}, this is, measure the difference of the specters of synthesized and real data. The C objective function is called spectral classification loss and it is defined as:

\begin{equation}
\begin{aligned}
    \mathcal{L}_{spectral} = \mathbb{E}_{x\sim p_{data}(x)} [log C(\phi(x))] \\
    + \mathbb{E}_{x\sim p_{g}(x)} [log (1 - C(\phi(x)))]
\end{aligned}
\end{equation}

\begin{figure}[h]
    \centering
    \includegraphics[width=\textwidth]{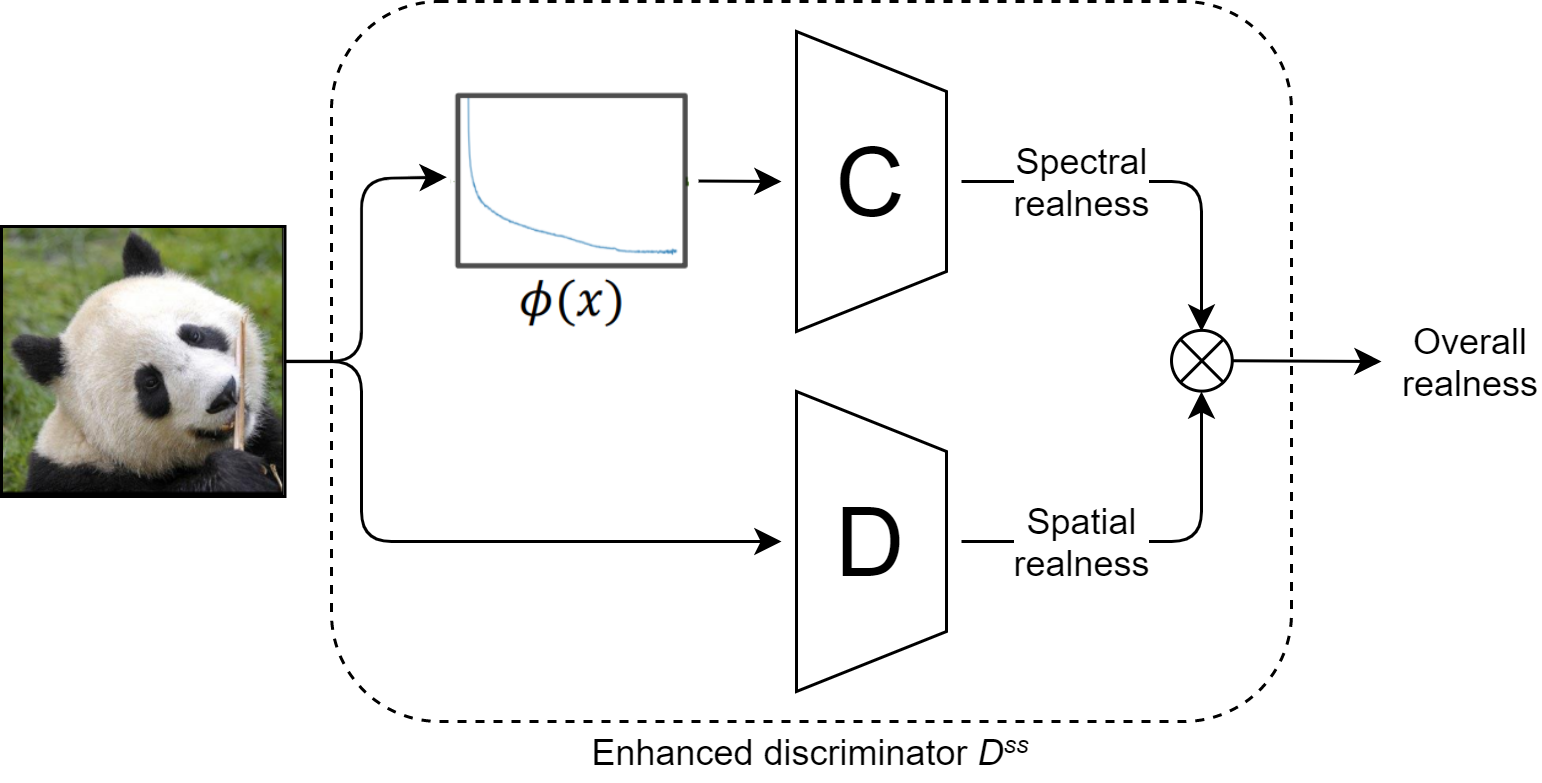}
    \caption{Structure of the proposed enhanced D of the \gls{ssdgan}. Figure based on Reference \cite{chen2020ssd}.}
\label{figure:SSD-GANDiscriminator}
\end{figure}

One of the strengths of the \gls{ssdgan} is its simplicity, easing implementation and allowing its implementation on various network architectures without excessive cost. The Fig. \ref{figure:SSD-GANDiscriminator} shows how both spatial and spectral information are processed by the new D proposed for the \gls{ssdgan}.

\gls{ssdgan} results show the potential of the proposed architecture. The quality of the images enhances the results of previous architectures, e.g. reducing the \gls{fid} score of the \gls{stylegan}~\cite{karras2019stylebased} from 4.40 to 4.06 by including the spectral classification.

\subsubsection{\gls{miegan}}
\label{section:miegan}
The \gls{miegan}~\cite{pan2021miegan} presents a novel architecture that aims to improve the quality of images taken with a mobile phone. To do so, two new networks are proposed, the so-called multi-mode cascade generative network and the adaptive multi-scale discriminative network. The generative network is composed of an Autoencoder architecture. The encoder of this new generator is divided into two streams, the inclusion of the second encoder is in charge of improving the low luminance areas, where mobile phones particularly lack in their clarity.

The discriminator network has a dual goal. First, the global discriminator ensures overall image quality. Second, a local discriminator maintains the local quality of small areas of the image. To combine both objectives, an adaptative weight allocation module is also proposed that is responsible for balancing the importance of each discriminator.

\begin{figure}[h]
    \centering
    \includegraphics[width=\textwidth]{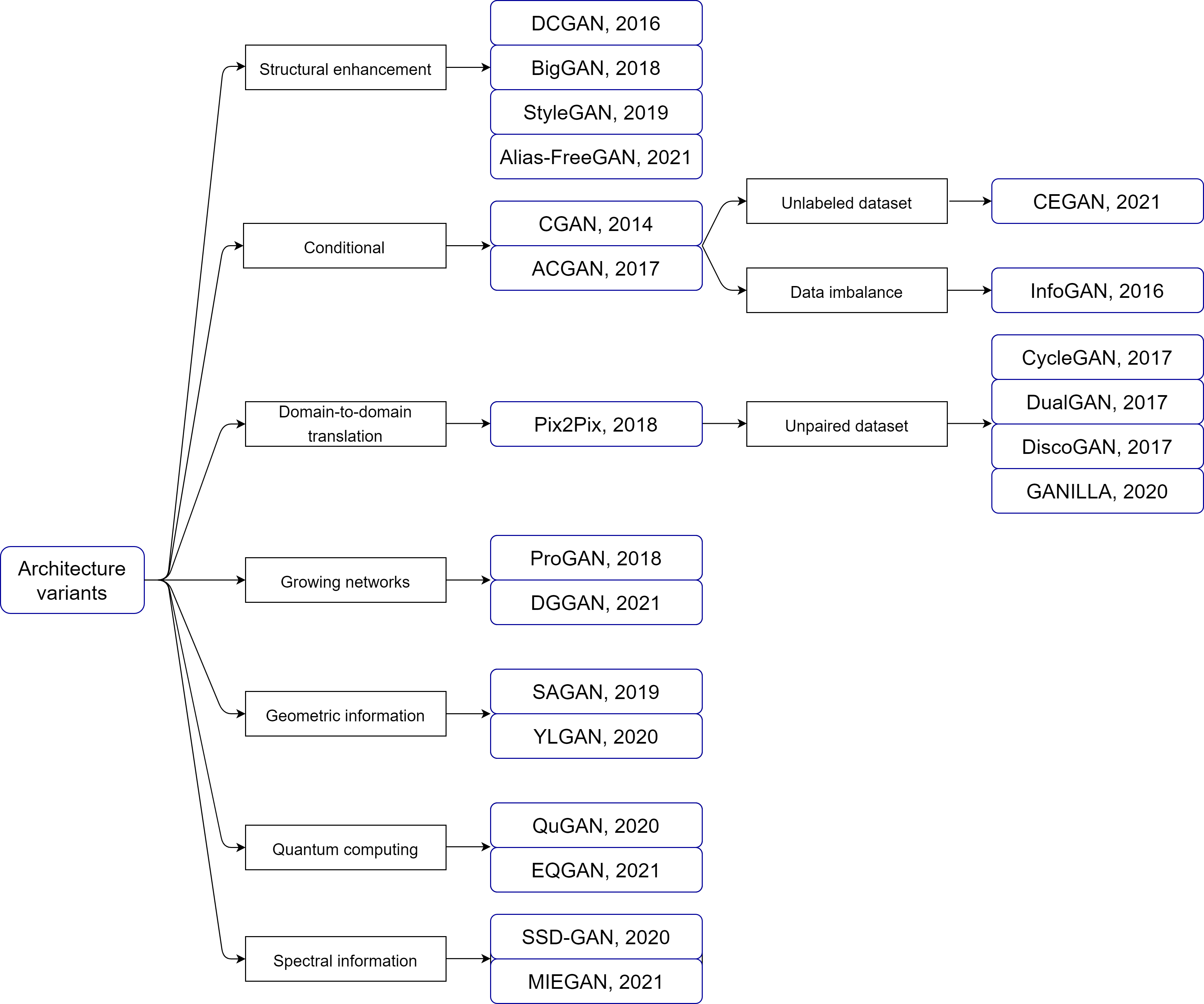}
    \caption{Survey proposed division of architecture variants for \glspl{gan}.}
\label{figure:ArchitectureVariants}
\end{figure}

A brief scheme reviewing all presented architecture variant \glspl{gan} can be seen in Fig.\ref{figure:ArchitectureVariants}. We divide the different architecture-based \glspl{gan} in different groups based on the proposed changes. The illustration gives a global view of how are interconnected different researches of the last years.

\subsection{Loss function optimization}
\label{section:LossFunctionOptimisationBasedGANs}

Orthogonal to the architecture modification \glspl{gan}, there are many researches\cite{arjovsky2017wasserstein, qi2017loss, Mao_2017_ICCV} that focuses on the objective function of \glspl{gan}. For example, the instability problem of \glspl{gan} is actually caused by the Jensen–Shannon divergence, where D often wins over G. Along with architecture optimization \glspl{gan}, there have been developed loss optimization researches, where both approaches coexist and interact with each other.

In this section, we will review the different most important and recent progress in variations of the loss function of the \glspl{gan}. 

\subsubsection{\gls{wgan}}
\label{section:wgan}
The base of the~\gls{wgan}~\cite{arjovsky2017wasserstein} is the application of the \textit{Earth Mover} (EM) distance, also known as \textit{Wasserstein-1} distance. The Wasserstein distance is defined as:
\begin{equation}
\begin{aligned}
    W(\mathbb{P}_{r}, \mathbb{P}_{g}) = \inf_{\gamma\in\Pi(\mathbb{P}_{r}, \mathbb{P}_{g})} \mathbb{E}_{(x, y)\sim\gamma} [||x-y||]
\end{aligned}
\end{equation}

In other words, the Wasserstein distance calculates the cost of transforming the distribution $ \mathbb{P}_{r} $ to the distribution $ \mathbb{P}_{g} $. In the case of GAN, the Wasserstein distance will measure the difference between the real and synthesized data distributions.

In order to apply the new objective function, some changes must be applied to the architecture of GANs. The D of the GAN changes its objective, but previously D was used to distinguish which data was real and which was synthesized in WGAN D change its name to \textit{critic}. The critic function is to measure the \textit{realness} of an image, e.g. the probability that the image belongs to the real distribution. The weight change of the critic is fixed between a window (e.g. between [-0.01, 0,01]) after each gradient update. The weight clipping is done to make the parameters lie in a compact space, due to the change of the critic network.

The EM distance has shown to produce better gradient behavior than other metrics. The results of the original paper show that, compared with the classical GAN loss function, the WGAN has better behavior in terms of convergence, mode collapse avoiding and stability. Particularly in low-dimensional manifold distributions, WGAN has shown to outperform traditional JS and KL divergences\cite{weng2019gan}. Other important benefit of WGAN is that the loss correlates with the quality of the synthesized samples and converges to a minimum.

WGAN is one of the most adopted variants, due to its capacity to deal with instability and mode collapse. Many later GAN variants\cite{rusu2016progressive, karras2019stylebased} use the WGAN loss function along with their own changes. For example, the \gls{mwgan}~\cite{cao2019multi} proposes a new objective function based on \gls{wgan} for multi marginal domain translation.

\subsubsection{\gls{wgangp}}
\label{section:wgangp}
In the original paper of \gls{wgan}, the authors suggest that weight clipping is "a terrible way to enforce Lipschitz constraints". Weigh clipping is one problem that the original \gls{wgan} had, but it worked well enough and its implementation was easy. The \gls{wgangp}~\cite{gulrajani2017improved} proposes a new technique to substitute the weight clipping that leads to the \gls{wgan} with undesired behavior.

The proposed change involves constraining the critic gradient norm output regarding to the input of the network. The constraint is softened via a penalty on the gradient norm. say that the new loss function is denoted as follows:
\begin{equation}
\begin{aligned}
    L = \mathbb{E}_{\overset{\sim}{x} \mathbb{P}_{g}} [D(\overset{\sim}{x})] - \mathbb{E}_{x \sim \mathbb{P}_{r}} [D(x)] + \\
    \lambda  \mathbb{E}_{\overset{\wedge}{x} \sim \mathbb{P}_{\overset{\wedge}{x}}} [(||\nabla_{\overset{\wedge}{x}} D(\overset{\wedge}{x})||_{2} -1)^{2}]
\end{aligned}
\end{equation}

The new change makes the \gls{wgangp} optimize its training, stabilizing it with almost no hyperparameter tuning. The new loss function also improves the quality of the generated images over \gls{wgan} and converges faster.

\subsubsection{\gls{ls-gan}}
\label{section:ls-gan}
In order to measure the quality of the synthesized samples of data created by G, a new loss function is used in the \gls{ls-gan}\cite{qi2017loss}. The new loss function aims to use regularization theory to improve the performance of GANs architecture. 
The main idea behind the new loss function is that a real sample produces smaller losses than a synthesized one, the margin between both is predefined. Once this assumption is set, we can infer that the training of G must aim at minimizing the loss margin between real and synthesized images. The proposed loss function is denoted as follows:
\begin{equation}
\begin{aligned}
    \min_{D}\mathcal{L}_{D} = \mathbb{E}_{x \sim p_{r}} L_{\theta} (x) + \lambda \mathbb{E}_{\underset{z \sim p_{z}}{x \sim p_{r}}} (\Delta(x, G(z))\\
    + L_{\theta} (x) + L_{\theta} (G(z)))_{+}
\end{aligned}
\end{equation}
\begin{equation}
\begin{aligned}
    \min_{G}\mathcal{L}_{G} = \mathbb{E}_{z \sim p_{z}} L_{\theta} (G(z))
\end{aligned}
\end{equation}
where $ \lambda $ is a hyperparameter for balancing and $ \theta $ are the parameters of D.

The loss function is regularized via Lipschitz regularity condition over the density of the real data. Due to the regularization, the created models are better in generalization of new data.

\subsubsection{\gls{lsgan}}
\label{section:lsgan}
The new loss function presented in \gls{lsgan}~\cite{Mao_2017_ICCV} aims to reduce the vanishing gradient problem. The main objective of the \gls{lsgan} is to punish the synthesized samples that are far from the real data but still in the correct side of the decision boundary. The least squares loss function is denoted as follows:
\begin{equation}
\begin{aligned}
    \min_{D} V_{LSGAN}(D) = \frac{1}{2} \mathbb{E}_{x \sim p_{data} (x)} [(D(x) - b) ^{2}] \\
    + \frac{1}{2} \mathbb{E}_{z \sim p_{z} (z)} [(D(G(z)) - a) ^{2}]
\end{aligned}
\end{equation}

\begin{equation}
\begin{aligned}
    \min_{G} V_{LSGAN}(G) = \frac{1}{2} \mathbb{E}_{z \sim p_{z} (z)} [(D(G(z)) - c) ^{2}]
\end{aligned}
\end{equation}
where $ a $ and $ b $ are the labels for fake and real data respectively and $ c $ is the label that G wants D to believe is real data. It should be noted that the square of both equations is responsible for punishing far from the decision boundary samples.

\begin{figure}[h]
    \centering
    
    \begin{subfigure}{0.49\columnwidth}
		\centering
		\includegraphics[width=\textwidth]{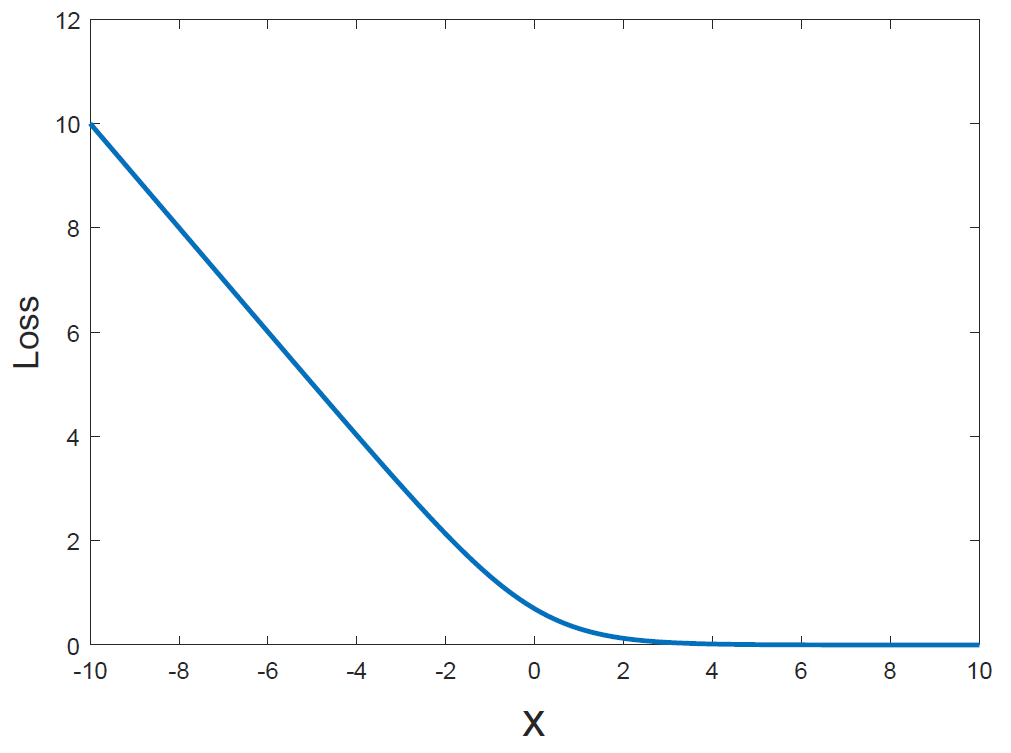}
		\caption{}
		\label{figure:SigmoidLossFunction}
	\end{subfigure}
	\hfill
	\begin{subfigure}{0.49\columnwidth}
		\centering
		\includegraphics[width=\textwidth]{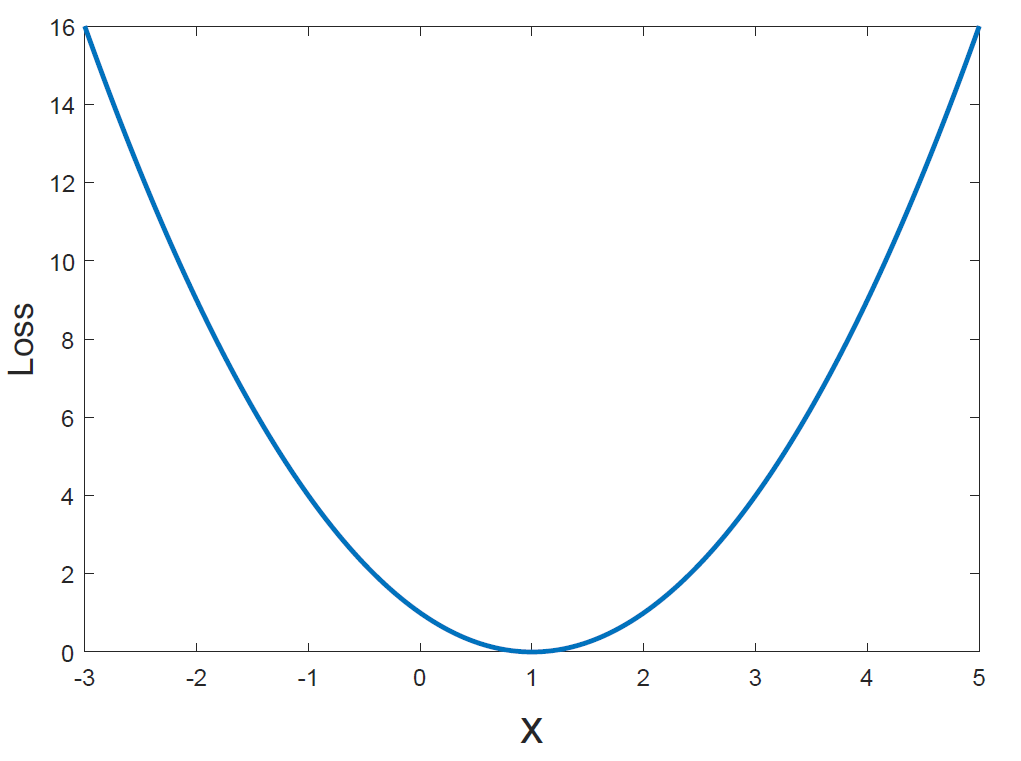}
		\caption{}
		\label{figure:LeastSquaresLossFunction}
	\end{subfigure}
	
    \caption{Comparison between sigmoid cross entropy loss function (\subref{figure:SigmoidLossFunction}) and least squares loss function (\subref{figure:LeastSquaresLossFunction}). Figure from Reference \cite{Mao_2017_ICCV}.}
    \label{figure:LSGANLosses}
\end{figure}

The \gls{lsgan} tries to generate more gradients while penalizing samples that lie a long way from the decision boundary. This way the gradients are forced to be higher, preventing the gradient vanishing problem. Compared to the classical sigmoid cross entropy loss function of GANs, the new least squares loss is flat only at one point as we can see in Fig.\ref{figure:LSGANLosses}.

\subsubsection{\gls{ugan}}
\label{section:ugan}
The \gls{ugan}~\cite{metz2017unrolled} loss function is defined to prevent instability in GANs training. The idea behind \gls{ugan} is to dynamically adapt G and D to prevent the situation of unbalance, where one of the networks is more trained than the other. Commonly, due to the particularity of the problem to solve, the D problem is easier to solve than the G one, producing an imbalance in favor of the D.

The training of \gls{ugan} is dynamically changed, the presented loss is surrogated for training the G. The surrogate objective function is created by unrolling $ K $ steps of D for each update of the G. Using the proposed loss function, the G behavior adapts to the training state of the D. The surrogate loss function is defined as follows:
\begin{equation}
\begin{aligned}
    \frac{ d f_{K}(\theta_{G}, \theta_{D})}{d\theta_{G}} = \frac{\partial f(\theta_{G}, \theta_{D}^{K}(\theta_{G},\theta_{D}))}{\partial\theta_{G}} \\
    + \frac{\partial f(\theta_{G}, \theta_{D}^{K}(\theta_{G},\theta_{D}))}{\partial\theta_{D}^{K}(\theta_{G}, \theta_{D})} \frac{d\theta_{D}^{K}(\theta_{G}, \theta_{D})}{d\theta_{G}}
\end{aligned}
\end{equation}

With the application of the proposed loss function, the \gls{ugan} demonstrates to stabilize the training by adjusting and synchronizing G and D networks. Furthermore, it prevents mode collapse, avoiding the model to drop regions of the data distribution. Despite this, the most important weakness of the \gls{ugan} is its computational cost. When the generator loss is optimized, the performance of the network drops. It depends on the particular problem how many unrolls need to stabilize its training. In the original paper, for example, it varies between 1 and 10.

\subsubsection{Realness GAN}
\label{section:RealnessGAN}
The new variation presented by RealnessGAN~\cite{xiangli2020real} is a generalization of the original version of the GAN. The proposed loss function changes the output of D, making it a distribution of the realness of the input data. In other words, the discriminator function is to measure the potential realness of the input data. The proposed loss function is defined as follows:
\begin{equation}
\begin{aligned}
    \max_{G}\min_{D} V(G, D) = \mathbb{E}_{x\sim p_{data}} [\mathcal{D}_{KL} (\mathcal{A}_{1}||D(x))] + \\
    \mathbb{E}_{x\sim p_{g}} [\mathcal{D}_{KL} (\mathcal{A}_{0}||D(x))]
\end{aligned}
\end{equation}
where $ \mathcal{A}_{0} $ and $ \mathcal{A}_{1} $ are the fake and real distributions.

Using the new loss function, the RealnessGAN is capable of recovering more modes than a standard GAN, preventing mode collapse. Furthermore, RealnessGAN shows a better performance, generating higher quality images in both real-world and synthetic datasets.

One of the strengths of the RealnessGAN is its simple implementation, due to the fact that RealnessGAN is a generalization of the original GAN. That said, despite being one of the most recently proposed architectures, it is expected to be widely used due to its good results and easy implementation.

\subsubsection{\gls{sngan}}
\label{section:sngan}
\gls{sngan}~\cite{miyato2018spectral} proposes a new technique to normalize the weights of D networks. A more stable training is searched through spectral normalization.

Respect previous normalizations\cite{salimans2016weight} spectral normalization is easier to implement. The previous methods imposed a much stronger constraint on the network matrix. With the spectral normalization, it is possible to relax this constraint, allowing the network to satisfy the local 1-Lipschitz constraint. The spectral normalization is defined as follows:
\begin{equation}
\begin{aligned}
    \bar{W}_{SN} := W/\sigma(W)
\end{aligned}
\end{equation}
where $ W $ is the weight matrix of D and $ \sigma(W) $ is the $ L_{2} $ normalization of W.

As mentioned before, the proposed D network is very simple and additionally its computational cost is small. It also requires the tuning of one hyperparameter, the Lipschitz constant.

The generated images using SN-GAN are more diverse, achieving better comparative IS respecting other weight normalizations.

\subsubsection{\gls{csgan}}
\label{section:csgan}
\gls{csgan}~\cite{kancharagunta2019csgan} proposes a new loss function for image-to-image translation problems. Previous works developed architectures for concrete domains of translation, \gls{csgan} proposes a common framework for different domain translation.

The Cyclic-Synthesized Loss (CS) is proposed as the objective function of \gls{csgan}. The new loss objective is to evaluate the differences between a synthesized image and its correspondent cycled image. The proposed loss function is denoted as follows:
\begin{equation}
\begin{aligned}
    \mathcal{L}(G_{AB}, G_{BA}, D_{A}, D_{B}) = \mathcal{L}_{LSGAN_{A}} + \mathcal{L}_{LSGAN_{B}}\\
    + \lambda_{A}\mathcal{L}_{cyc_{A}} + \lambda_{B}\mathcal{L}_{cyc_{B}} + \mu_{A}\mathcal{L}_{CS_{A}} + \mu_{B}\mathcal{L}_{CS_{B}}
\end{aligned}
\end{equation}
were $ \mathcal{L}_{CS_{A}} $ and $ \mathcal{L}_{CS_{B}} $ are the Cyclic-Synthesized loss of both domains.

With respect to previous architectures, \gls{csgan} produces images of better quality, notably reducing the artifacts of the synthesized images. The results show better performance of \gls{csgan} in \gls{cuhk} dataset~\cite{wang2008face} and comparable performance in FACADES dataset~\cite{tylevcek2013spatial}. The comparison of the performance is made against \gls{gan}~\cite{goodfellow2014}, \gls{pix2pix}~\cite{isola2018imagetoimage}, \gls{dualgan}~\cite{yi2017dualgan}, \gls{cyclegan}~\cite{zhu2017unpaired} and \gls{ps2man}~\cite{wang2018high}.

\subsubsection{\gls{missgan}}
\label{section:missgan}
The proposed architecture of \gls{missgan}~\cite{barzilay2021miss} presents only one trained model to generate illustrations for different image styles. Previous methods used different G for each style, limiting the practical application of the architectures, while \gls{missgan} uses a unique model.

\begin{figure}[h]
    \centering
    \includegraphics[width=\textwidth]{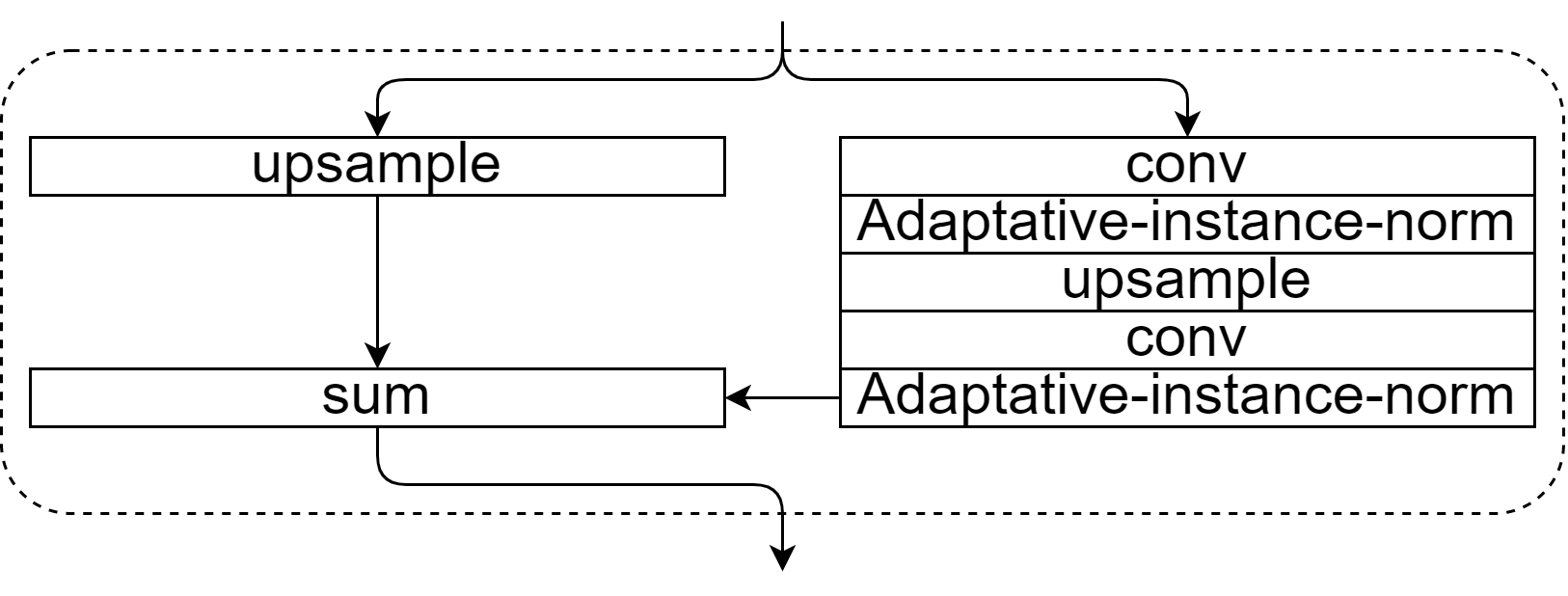}
    \caption{Structure of the proposed residual blocks of the \gls{missgan}. Figure based on Reference \cite{barzilay2021miss}.}
\label{figure:MISSGANResidual}
\end{figure}

The proposed new G is based on the GANILLA~\cite{hicsonmez2020ganilla} architecture, but it proposes some changes to the architecture of the decoder of the GANILLA G. The new decoder contains three residual blocks, these residual blocks are in charge of processing the low-level features from previous layers. The composition of each residual block can be seen in Fig.\ref{figure:MISSGANResidual}.

To train the \gls{missgan} models five different objective functions are proposed.

The first loss function is called the \textit{adversarial objective} ($ \mathcal{L}_{adv} $) and it is in charge of, taking the input image and the target domain, ensure that the generated image style corresponds with the target domain. To do so the $ \mathcal{L}_{adv} $ takes two discriminator predictions, one for the input image and other for the synthesized image.

The second loss function is denoted as \textit{style reconstruction objective} ($ \mathcal{L}_{sty} $), and it enforces the G to use the mapping network style code while receiving a generated latent code, to calculate the $ \mathcal{L}_{sty} $ the output of the G encoder over the generated image.

The third proposed objective function is called \textit{style diversification objective} ($ \mathcal{L}_{ds} $) and it compares a pair of synthesized images, each image corresponds to a different style code, each one generated from a different latent code. The objective of this loss function is to force G to produce diverse images, preventing two images with different latent codes from being the same.

The fourth objective function is the \textit{cycle consistency loss} ($ \mathcal{L}_{cyc} $) used in the \gls{cyclegan}~\cite{zhu2017unpaired}.

Finally, the fifth objective function is called \textit{content features loss} ($ \mathcal{L}_{content\_feat} $), and it computes the distance in the feature space by using a VGG16~\cite{simonyan2015verydeep} network.

To combine the different objective function a \textit{total objective} is defined as follows:

\begin{equation}
\begin{aligned}
    \max_{D}\min_{G, F, E} \mathcal{L}_{adv} + \lambda_{sty}\mathcal{L}_{sty} - \lambda_{ds}\mathcal{L}_{ds} \\
    + \lambda_{cyc}\mathcal{L}_{cyc}
    + \lambda_{feat}\mathcal{L}_{content\_feat}
\end{aligned}
\end{equation}

where $ E $ is the style encoder and $ F $ is the mapping network, all the $ \lambda $ parameters correspond to a hyperparameter for each objective function.

\subsubsection{Sphere GAN}
\label{section:SphereGAN}
SphereGAN~\cite{park2019sphere} proposes a new architecture based on integral probability metric (IPM). The main characteristic of SphereGAN is that it bounds the IPMs objective function on a hypersphere.

Compared with other architectures such as WGAN-GP~\cite{gulrajani2017improved} SphereGAN loss function does not require any constraint term, reducing the necessity of hyperparameter tuning. The loss function of SphereGAN is defined as follows:

\begin{equation}
\begin{aligned}
    \min_{G}\max_{D}\sum_{r} E_{x} [d_{s}^{r} (N, D(x))] - \sum_{r} E_{z} [d_{s}^{r} (N, D(G(z)))]
\end{aligned}
\end{equation}

where $ d_{s}^{r} $ denotes the $ r $-th moment distance between a sample and the north pole of the hypersphere.

In the original paper, the mathematical properties of SphereGAN are proved, showing that minimizing the objective function of SphereGAN is equivalent to reducing IPM. In addition, it is proved that SphereGAN compared to WGAN can use r-Wasserstein distances, unlike WGAN that could only use 1-Wasserstein distance. This provides to SphereGAN a wider function space.

The SphereGAN results show its good performance, achieving a IS of 8.39 and \gls{fid} score of 17.1 in CIFAR-10~\cite{krizhevsky2009learning} dataset. Compared to \gls{wgangp} that achieved IS of 7.86 in the same dataset.

\subsubsection{\gls{srgan}}
\label{section:srgan}
In order to apply \glspl{gan} to image upscaling the \gls{srgan}~\cite{ledig2017photo} was proposed. The proposed \gls{gan} objective is to take an input natural image and upscale it resolution by a factor of 4.

To achieve the super resolution, the new variant proposes a couple of adversarial and content losses. Both functions are combined using the called \textit{perceptual loss function}, this function is in charge of ass solution respecting the relevant characteristics of the data. The content loss is defined as follows:

\begin{equation}
\begin{aligned}
    l^{SR} = l_{X}^{SR} + 10^{-3}l_{Gen}^{SR}l_{Gen}^{SR}
\end{aligned}
\end{equation}

where $ l_{Gen}^{SR} $ is the adversarial loss and $ l_{X}^{SR} $ is the content loss.

The content loss used relies on a pre-trained VGG-19 model~\cite{simonyan2015verydeep}. This model, respecting the usage of a loss function such as \gls{mse} is more invariant to changes in pixel space. This metric will provide the network information about the quality of the content of the synthesized image. The new loss function is calculated as:

\begin{equation}
\begin{aligned}
    l^{SR}_{VGG/i,j} = \frac{1}{W_{i,j} H_{i,j}} \sum_{x=1}^{W_{i,j}} \sum_{y=1}^{H_{i,j}} ( \phi_{i,j}(I^{HR})_{x,y} \\
    - \phi_{i,j}(G_{\theta G} (I^{LR}))_{x,y})^{2}
\end{aligned}
\end{equation}

where $ I^{LR} $ refers to the low resolution images and $ I^{HR} $ refers to the high resolution image.

In addition to the content loss, the adversarial loss is defined as being this part of the generative component of the \gls{gan}. This function is responsible for pushing the generated images to be realistic and indistinguishable from the real ones. The loss function is defined as:

\begin{equation}
\begin{aligned}
    l^{SR}_{Gen} = \sum_{n=1}^{N} - log D_{\theta D} (D_{\theta G} (I^{LR}))
\end{aligned}
\end{equation}

The application of \gls{srgan} improves the results of previous algorithms for image super resolution.

Since the introduction of the \gls{srgan} it has been used in many different applications~\cite{zhang2021super, dehzangi2021oct, zhao2021deep}. In addition, there are works such as \cite{liu2021super} that presents some improvements in the \gls{srgan} structure, the new architecture is known as \gls{srcagan}. The architecture presented in this papers adds a channel attention module to the models, this module recovers the attention layer used in \gls{sagan}~\cite{zhang2019self}. The results presented in this new architecture outperforms the \gls{srgan}.

\subsubsection{\gls{wsrgan}}
\label{section:wsrgan}
One of the characteristics of the \gls{srgan}~\cite{ledig2017photo} was the combination of the content loss and adversarial loss during the training. The \gls{wsrgan} proposes is changing the importance of each loss and studying the effect of this action.

The main objective of the \gls{wsrgan} is to improve the performance of the architecture by analyzing its performance in different combinations of its objective functions. Then the new weighted loss function is defined as follows:

\begin{equation}
\begin{aligned}
    l^{SR}_{X} = w l_{MSE}^{SR} + (1-w) 10^{-3}+l_{VGG}^{SR}
\end{aligned}
\end{equation}

where $ w $ is the parameter that controls the impact of each loss function on the final result.

After training the network with different weight configurations, the paper concludes that the \gls{mse} loss is the most important loss function, being supported by the VGG loss.

Additionally, the definition of the weight parameter is declared dynamically, obtaining even better results than when it is static.

\begin{figure}[!h]
    \centering
    \includegraphics[width=\textwidth]{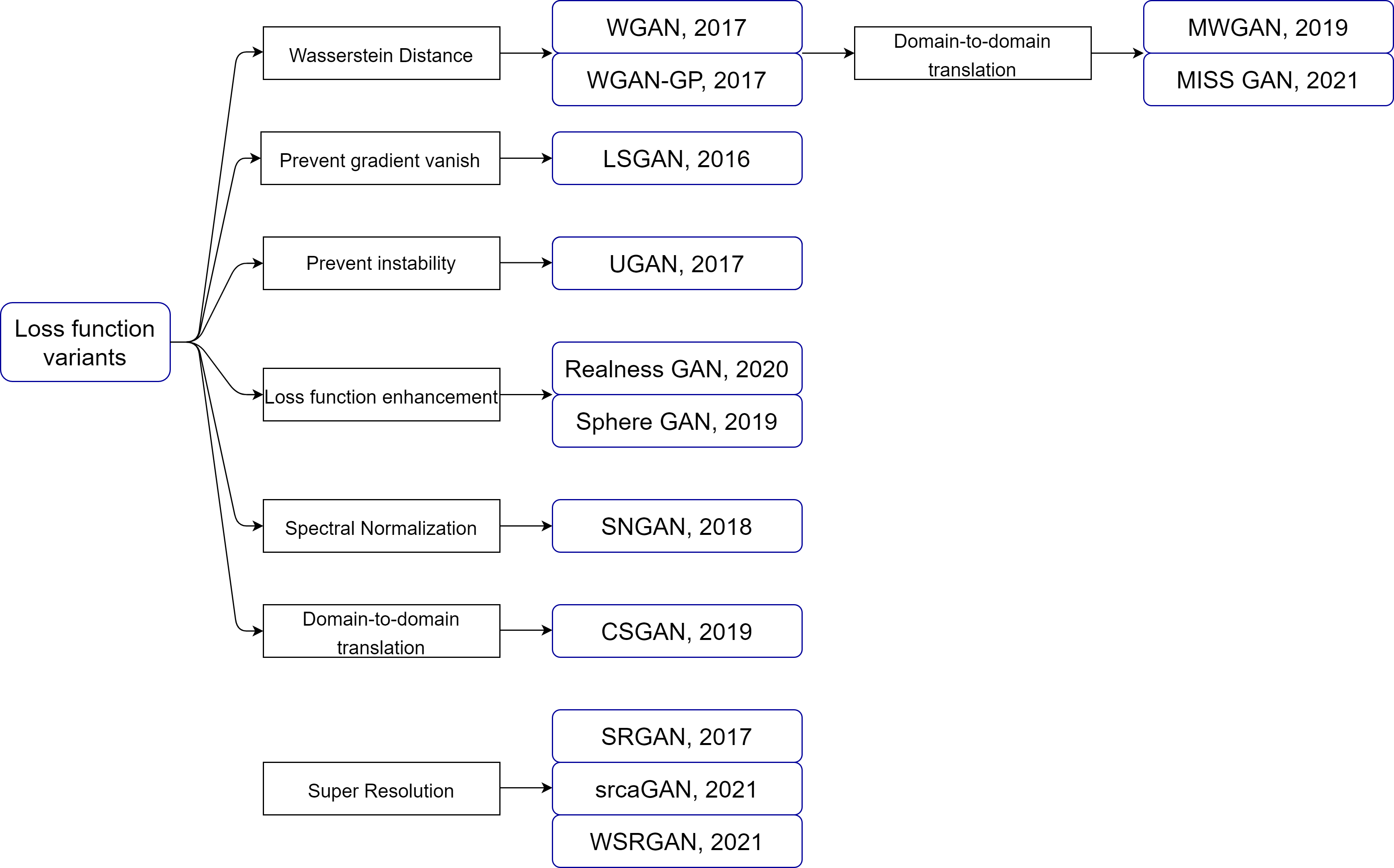}
    \caption{Survey proposed division of loss function variants for \glspl{gan}.}
\label{figure:LossVariants}
\end{figure}

A brief scheme reviewing the different presented loss function variant \glspl{gan} can be seen in Fig.\ref{figure:LossVariants}. We divide the different \glspl{gan} in different groups based on the proposed changes in the loss function.

\subsection{\gls{gan} timeline}
\label{section:GANTimeline}
\begin{figure}[!h]
    \centering
    \includegraphics[width=\textwidth]{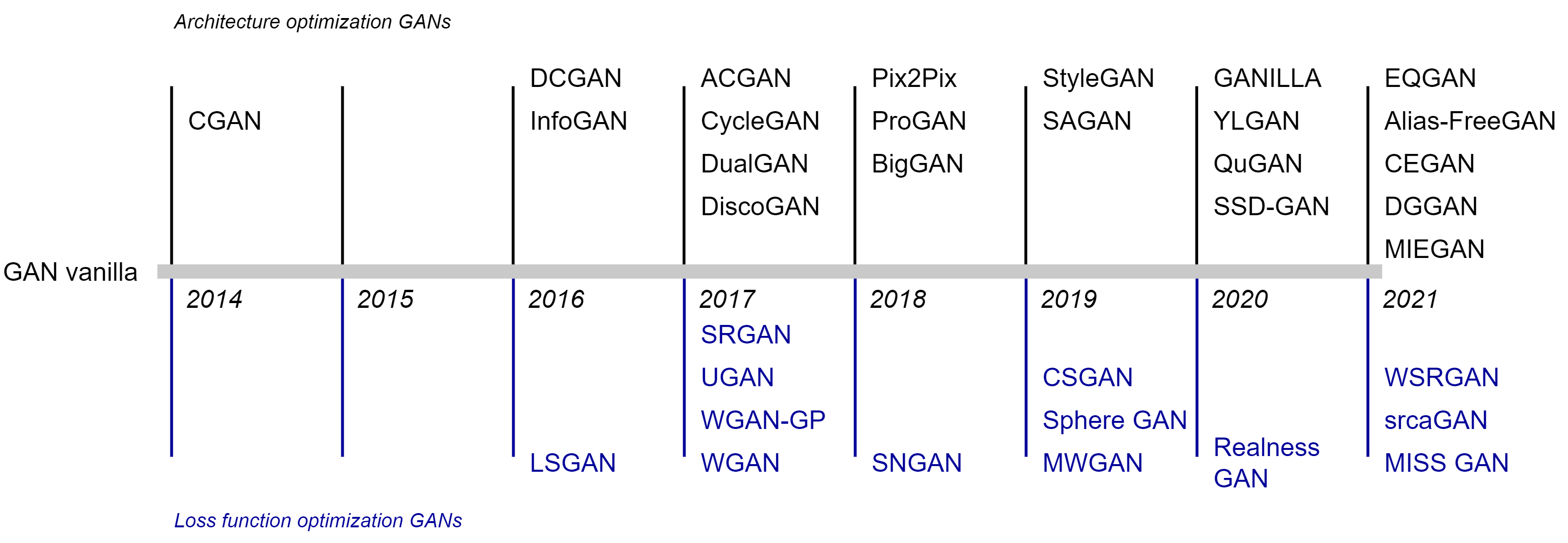}
    \caption{Timeline of the reviewed GAN architectures.}
\label{figure:GANTimeline}
\end{figure}

A timeline with the reviewed architectures is presented in figure \ref{figure:GANTimeline}. The \glspl{gan} that have been studied during sections \ref{section:ArchitectureOptimisationBasedGANs} and \ref{section:LossFunctionOptimisationBasedGANs} are showed temporally. This timeline provides an overview of the historical development of \glspl{gan}.

As it can be seen, the timeline compiles the most important works of the last decade. It is important to analyze that some researches have influenced posterior ones. In some cases some researches adopt innovation of previous works as a base to then propose new changes, e.g. the \gls{dcgan} that have influenced several posterior works. In other cases there are relationships between works can be seen as an unique research, linking each article with each other by taking previous results and improving them, e.g. in the case of \gls{progan}, \gls{stylegan} and Alias-Free GAN.

\section{GAN applications}
\label{section:GANApplications}
As mentioned before,~\glspl{gan} are one of the most popular applications of machine learning of the last years. \glspl{gan}~models can achieve results in fields where previous models could not, in other cases, \glspl{gan} improve the previous results significantly.

In this section, we will review the most important fields where \gls{gan}~architectures are applied, paying a special attention to the \gls{gan} models related to computer vision tasks and we will compare the different architecture results.

Most of the last researches focus on how to apply \glspl{gan}~to generate new synthesized data, replicating a data distribution. But, as we will review in this section \glspl{gan} can be applied to other fields, \eg video game creation~\cite{kim2020learning}.

\subsection{Image synthesis}
\label{section:ImageSynthesis}
One of the most important fields in which~\glspl{gan} are applied is in computer vision. In particular, realistic image generation is the most widely used application of \glspl{gan}~\cite{karras2019stylebased, karras2018progressive, arjovsky2017wasserstein}.

Most of the proposed \gls{gan} variants are tested by generating real world images. Arguably, image synthesis is the first application one might think of when thinking about \gls{gan}. Its popularity is due to the good results that \gls{gan} can achieve. Compared with previous methods, \glspl{gan} provide sharper results~\cite{genevay2017gan}. Both in academic world and for the general public \gls{gan} has raised a lot of interest.

One of the main reasons of the \gls{gan} success is its results easy understanding. As the mainly generated output of \glspl{gan} are images, they can be easily understood by anyone. Even if a person does not have any technical understanding of artificial intelligence, it is possible to judge the results.

Within computer vision, image generation is the most used method to test \glspl{gan}. There are plenty of real world images datasets that can be used to train \glspl{gan}. The availability of datasets that can be used for training neural networks is usually the main drawback of artificial intelligence projects. Either by its availability or by its content~\cite{denton2020bringing} having a good dataset is essential for machine learning. When real world images are used to train \gls{gan} models, the availability of good datasets is not a problem, there are a large variety of datasets~\cite{deng2009imagenet, krizhevsky2009learning} that have been widely tested and are well known in the academic community.

Since the first~\gls{gan} publication~\cite{goodfellow2014}~\gls{gan} architectures have been used for synthesizing real world images. In the original proposed \gls{gan} the models were used to generate images replicating MNIST~\cite{lecun1998gradient}, CIFAR-10~\cite{krizhevsky2009learning} and \gls{tfd}~\cite{susskind2010toronto} datasets. The generated images using the original structure were very blurry and did not have good quality. Besides that, the presented results supposed the presentation of the \gls{gan} architecture.

One of the first improvements to the original architecture was the \gls{dcgan}~\cite{radford2016unsupervised}, it proposed structural changes and hyperparameter tunning respect the first proposed model. The results of the \gls{dcgan} showed improvements in the performance and generation of the networks, the generated images were clearer and more recognizable. Despite that, the architecture still suffered from instability and mode collapse.

The \gls{wgan} architecture~\cite{arjovsky2017wasserstein} could reduce drastically the mode collapse and instability of the previous models. Thus, later models adapted the loss function of the \gls{wgan} along with their respective structural changes in the network.

Recently the \gls{progan}~\cite{karras2018progressive} introduced a new training methodology that achieved an improved performance of the networks. With the new methodology came a huge improvement in the quality of the generated images. The results showed not only a more stable trainings but sharper, with finer details and more diverse images. Due to the particularities of the applied methodology, it can be applied to other architectures, so in later works the \gls{progan} training methodology will be used as its base.

Following the line of research of \gls{progan} the \gls{stylegan}~\cite{karras2019stylebased} was presented. The results produced by the \gls{stylegan} could improve the results of the \gls{progan}. At this point some generated datasets, \eg human faces images, were indistinguishable from real images from a human perception. Along with the high quality of the images the \gls{stylegan} proposed style mixing, capable of generating new images combining previous images. This allows to modify image features at a high, medium and low level, allowing the network to disentangle different features of an image, providing more control of the generated images.

One of the main problems of the \gls{stylegan} was the known as \textit{texture sticking}. This caused the generated images to have a certain texture in an absolute position. When interpolating different images it was noticeable that some parts of the images, \eg the hair of a human face, maintain the same texture in spite of changing its position. The Alias-Free GAN proposed an architecture that suppressed the texture sticking problem. By eliminating the sticking problem, the interpolation of synthesized images is smoothed, generating a continuum of images, not only realistic individually but also as a set. The improvements of the Alias-Free GAN together with the style mixing of \gls{stylegan} allows to create animations of, for example, a human face changing its position, gender or features such as the smile.

The Table \ref{table:comparisonimagesynthesis} summarizes the performance of the presented \gls{gan} models during this section. The compared datasets are MNIST~\cite{lecun1998gradient}, \gls{tfd}~\cite{susskind2010toronto}, CIFAR-10~\cite{krizhevsky2009learning}, CelebA-HQ~\cite{karras2018progressive} and \gls{ffhq}~\cite{karras2019stylebased}. The used metric for comparing the different variants are accuracy of the models (the higher score the better ↑), \gls{is} (the higher score the better↑) and \gls{fid} (the lower score the better ↓)

\begin{table*}[htbp]\footnotesize
    \caption{Performance summary of image generation \glspl{gan}}
    \centering
    \begin{tabular}{|c||c|c||c||c|}
    \hline
    \multirow{2}{*}{Model} & \multicolumn{2}{c|}{CIFAR-10} & CelebA-HQ & \gls{ffhq} \\ \cline{2-5} 
     & Accuracy ↑ & \gls{is} ↑ & \gls{fid} ↓ & \gls{fid} ↓ \\ \hline
    \gls{dcgan} & \textbf{82.8\%} & 6.58 & - & - \\ \hline
    \gls{progan} & - & \textbf{8.80} & 7.79 & 8.04 \\ \hline
    \gls{stylegan} & - & - & \textbf{5.06} & 4.40 \\ \hline
    \gls{stylegan}2 & - & - & - & \textbf{2.70} \\ \hline
    Alias-Free GAN & - & - & - & 3.07 \\ \hline
    \end{tabular}
    \label{table:comparisonimagesynthesis}
\end{table*}

\subsection{Image-to-image translation}
\label{section:Image-to-imageTranslation}
Taking an image from one domain and converting it to the other domain is known as image-to-image translation. It was first proposed with the \gls{pix2pix} architecture~\cite{mirza2014conditional}, \gls{pix2pix} is based on \gls{cgan} following the idea of generating images conditioned on their composition via a label input. With \gls{pix2pix} the networks are capable of learning how the same image is translated between one domain and another. The main drawback that \gls{pix2pix} had was the requirement of having a paired dataset of images in both domains.

Following the steps of \gls{pix2pix} \gls{cyclegan}~\cite{zhu2017unpaired}, \gls{dualgan}~\cite{yi2017dualgan} and \gls{discogan}~\cite{kim2017learning} were developed. These new architectures were based on the cyclic consistency idea. Cyclic consistency was previously used in machine learning~\cite{sundaram2010dense, kalal2010forward}, it is based on the idea that translating an image from one domain to another and then doing the reverse operation will recover the original image. Following this concept the new networks were capable of translating images without a paired dataset. By not needing a paired dataset the number of possible applications of \gls{gan} to image-to-image translation increased considerably.

Later on the \gls{csgan} was proposed\cite{kancharagunta2019csgan} improving the results of previous architectures. The new proposed loss function achieved better results in image generation, comparing with \gls{cyclegan}~\cite{zhu2017unpaired}, \gls{dualgan}~\cite{yi2017dualgan}, \gls{discogan}~\cite{kim2017learning} and \gls{ps2man}~\cite{wang2018high}. This new architecture results follows the natural progression of the \gls{gan} in image-to-image translation and promise an exciting future in what \gls{gan} can do.

The image-to-image translation is especially popular in society, because of the applications that have been developed in the last years. With the architecture of the presented \glspl{gan} the general public is capable, for example, of taking a personal image of themselves and transforming it into one of an old person with his face. This type of applications have become popular in social networks, increasing their visibility even more.

This interaction between society and \gls{gan} development is mutually beneficial, the society uses the technological advances of the last years while the academic community gain impact and repercussion. From an academic perspective this interaction should be considered positive and it should be noted that most of the impact of machine learning during the last years have been caused by the publicity given by the mass media and the social networks. Although most of the people are not interested in the technique behind \gls{gan} applications they act as a catalyst to make more people interested in artificial intelligence and, ultimately, it will bring more people to academic research in the field.

The Table \ref{table:comparisonimagetoimagetranslation} summarizes the performance of the presented GAN models in image-to-image translation tasks. The data is obtained from~\cite{kancharagunta2019csgan}, where the \gls{ssim} (the higher score the better ↑), \gls{mse} (the lower score the better ↓), \gls{psnr} (the higher score the better ↑) and \gls{lpips}~\cite{zhang2018unreasonable, lin2018conditional} (the lower score the better ↓) are computed for different \gls{gan} variants. The comparison is made for \gls{cuhk}~\cite{wang2008face} and FACADES~\cite{tylevcek2013spatial} datasets. The \gls{lpips} is a metric that measure the distance between the real and the generated distribution via perceptual similarity.

\begin{table*}[htbp]\footnotesize
    \caption{Performance summary of image-to-image translation \glspl{gan}}
    \begin{center}
        \begin{tabular}{|c||c|c|c|c||c|c|c|c|}
        \hline
        \multirow{2}{*}{Model} & \multicolumn{4}{c|}{\gls{cuhk}} & \multicolumn{4}{c|}{FACADES} \\ \cline{2-9} 
         & \gls{ssim} ↑ & \gls{mse} ↓ & \gls{psnr} ↑ & \gls{lpips} ↓ & \gls{ssim} ↑ & \gls{mse} ↓ & \gls{psnr} ↑ & \gls{lpips} ↓ \\ \hline
        \gls{gan} & 0.5398 & 94.8815 & 28.3628 & 0.157 & 0.1378 & 103.8049 & 27.9706 & 0.525 \\ \hline
        Pix2Pix & 0.6056 & 89.9954 & 28.5989 & 0.154 & 0.2106 & \textbf{101.9864} & \textbf{28.0568} & \textbf{0.216} \\ \hline
        \gls{dualgan} & 0.6359 & 85.5418 & 28.8351 & 0.132 & 0.0324 & 105.0175 & 27.9187 & 0.259 \\ \hline
        \gls{cyclegan} & 0.6537 & 89.6019 & 28.6351 & 0.099 & 0.0678 & 104.3104 & 27.9489 & 0.248 \\ \hline
        \gls{ps2man} & 0.6409 & 86.7004 & 28.7779 & 0.098 & 0.1764 & 102.4183 & 28.032 & 0.221 \\ \hline
        \gls{csgan} & \textbf{0.6616} & \textbf{84.7971} & \textbf{28.8693} & \textbf{0.094} & \textbf{0.2183} & 103.7751 & 27.9715 & 0.22 \\ \hline
        \end{tabular}
        \label{table:comparisonimagetoimagetranslation}
    \end{center}
\end{table*}

\subsection{Video generation}
\label{section:VideoGeneration}
\glspl{gan} have proven to generate state-of-the-art results in image processing. Along with image generation comes the possibility to generate a set of images generating a video. Video generation is a more complex task than image generation. The issues associated with image generation are included in video generation, but the computational cost of training models that can process video is high. In addition, the synthesized videos must be coherent.

One of the particular problems of video is the motion blur generated by the networks\cite{guo2021exploring}. When a video is generated, the tracking of some objects can be difficult, generating fuzziness in some portions of the image. Some works have tried to tackle this problem\cite{zhang2020deblurring, younus2020effective, ren2020video}, but it is still an open problem.

One of the most popular applications of video generation with \glspl{gan} is the known as \textit{deep fake}. Deep fake consists in taking a video of a person and changing the face of the human to be someone else. Many works have been developed in the last years in this field\cite{westerlund2019emergence}.

Deep fake is one of the most controversial applications of \gls{gan}, the possibility of changing a face in a video allows to generate fake videos that can be used to supplant a person. This problem is magnified in the case of women\cite{martinez2019historia} due to their position in society. Even so, there are some applications of deep fake where it can be beneficial\cite{kwok2021deepfake}, its application still raises doubts in the society. This is why many recent researches have focused on how to detect deep fake videos\cite{korshunov2019vulnerability, dolhansky2020deepfake, carlini2020evading, zhao2021multi}.

Other application of \glspl{gan} to video generation are video-to-video translation, which is indeed the general case of deep fake. Many architectures of this type have been proposed during the last years\cite{chen2019mocycle, bansal2018recycle}.

It should be noted that, in the case of video processing, the standard is to use previous information, such as another video, to generate the synthesized data. Unlike image generation, video generation is more interesting if the new information is conditioned by an external agent. In image processing, the only input was the latent space, but the final images were conditioned by the dataset of the training. When videos are generated, the degree of freedom is extended, enabling the generated data to be less controlled. Controlling the video output is necessary to maintain the coherence of the final output, but it also eases the \gls{gan} job, which is significantly more difficult with respect to image processing.

\subsection{Image generation from text}
\label{section:ImageGenerationFromText}
Since the introduction of \gls{cgan} the capabilities of \glspl{gan} were expanded. The possibility of constraint the synthesized information that \glspl{gan} produced made the networks have a wider range of application. By controlling the output of the generations of the networks the applications of them can be much more specific and interesting. One field were \glspl{gan} have shown to outperform previous techniques in image generation from text~\cite{kurup2021evolution}.

\gls{stackgan}~\cite{zhang2017stackgan} was one of the firsts proposed architectures for image generation from text. The architecture splits in two stages, the generation problem, the objective is to divide the main problem in sub-problems that are easier to handle in the network. The known as \textit{Stage-I GAN} is in charge of producing a coarse sketch of the desired image, this way this part of the network focuses on translating the text to a image that fulfills the description. Then, the \textit{Stage-II GAN} takes the generated image from Stage-I GAN, increases its resolution and define the finer details. The \gls{stackgan} is able of producing images that match the input description while achieving sharp, high quality samples. Later on the \gls{stackgan++}~\cite{zhang2018stackgan++} was proposed, this new architecture resolved some problems of the original \gls{stackgan}, stabilizing its training and improving the overall quality of the synthesized images.

One problem of the \gls{stackgan} is that it is highly dependent on the sketch generated by the Stage-I GAN. To solve this \gls{dmgan} proposed a new technique based on memory networks~\cite{gulcehre2016dynamic, weston2014memory} that divides the generation problem in two steps. In the first one a initial image is generated and in the second step a memory network is used to refine the details and produce a high quality image. To connect the memory and the \gls{gan} a \textit{response gate} is proposed, by controlling dynamically the flow of information the gate is capable of fusing the information appropriately. The results of the \gls{stackgan} shows a higher quality respecting all previous architectures.

\gls{dualattentionalgan} proposed a new architecture based on two modules. The \gls{vam} is in charge of taking care of the internal representations of the image information, capturing the global structures and their relationships. The \gls{tam} defines the relations between the text and the image, defining the links between both. Finally a \gls{aem} fuse the visual with the textual information, concatenating them along with the input features of the image. The results of the \gls{dualattentionalgan} shows an improved performance respecting previously used architectures.

Following the general architecture of \gls{stackgan} \gls{dfgan} was proposed~\cite{tao2020df}. The \gls{dfgan} architecture only have one stage of image generation, this backbone synthesized new images conditioned by an input text using only one pair of G and D. Thus being a simpler structure, \gls{dfgan} achieves better performance and efficiency compared with previous variants. The new techniques that \gls{dfgan} proposes are a new fusion module, known as \textit{deep text-image fusion block}, and a new discriminator capable of promoting the generator to synthesize higher quality images without extra networks. The results of the \gls{dfgan} shows an improvement on the quality of the images, without committing to more complex models and improving the efficiency of the previous architectures.

The one-stream information approach followed in \gls{dfgan} was reused in \gls{ldcgan}~\cite{gao2021lightweight}. The proposed architecture of the \gls{ldcgan} consists on one G and two independent discriminators. The generator is composed by a \gls{ce} that disentangles the features of the input text by using unsupervised learning. Then a \gls{cmb} provides continuously the images features with the compensation information. Finally using the known as \gls{parb} the generated image is enriched maintaining multiscale context and spatial multiscale features. The results of the architecture not only shows a higher quality image respecting previous methods, but also improves the performance decreasing the number of parameters by 86.8\% and the computation time by 94.9\%.

The Table \ref{table:comparisonimagefromtext} summarizes the performance of the presented \gls{gan} models during this section. In addition to the mentioned networks the \gls{ganintcls}~\cite{reed2016generative} and the \gls{gawwn}~\cite{reed2016learning} are included, both of this networks act as a reference of previous architectures. The compared metrics are \gls{hr} (the lower score the better ↓), \gls{is} (the higher score the better ↑) and \gls{fid} (the lower score the better ↓). The compared datasets are \gls{coco}~\cite{lin2014microsoft}, \gls{cub}~\cite{wah2011caltech} and Oxford-102~\cite{nilsback2008automated}.

\begin{table*}[htbp]\footnotesize
    \caption{Performance summary of image generation from text \glspl{gan}}
    \centering
    \begin{tabular}{|c||c|c|c||c|c|c||c|c|c|}
    \hline
    \multirow{2}{*}{Model} & \multicolumn{3}{c|}{\gls{coco}} & \multicolumn{3}{c|}{\gls{cub}} & \multicolumn{3}{c|}{Oxford-102} \\ \cline{2-10} 
     & \gls{hr} ↓ & \gls{is} ↑ & \gls{fid} ↓ & \gls{hr} ↓ & \gls{is} ↑ & \gls{fid} ↓ & \gls{hr} ↓ & \gls{is} ↑ & \gls{fid} ↓ \\ \hline
    \gls{ganintcls} & 1.89 & 7.88 & - & 2.81 & 2.88 & - & 1.87 & 2.66 & - \\ \hline
    \gls{gawwn} & - & - & - & 1.99 & 3.62 & - & - & - & - \\ \hline
    \gls{stackgan} & \textbf{1.11} & 8.45 & - & 1.37 & 3.70 & - & \textbf{1.13} & 3.20 & - \\ \hline
    \gls{stackgan++} & 1.55 & 8.30 & 81.59 & \textbf{1.19} & 4.04 & 15.30 & 1.30 & 3.26 & 48.68 \\ \hline
    \gls{dmgan} & - & \textbf{30.49} & 32.64 & - & 4.75 & 16.09 & - & - & - \\ \hline
    \gls{dualattentionalgan} & - & - & - & - & 4.59 & \textbf{14.06} & - & \textbf{4.06} & \textbf{40.31} \\ \hline
    \gls{dfgan} & - & - & \textbf{21.42} & - & \textbf{5.10} & 14.81 & - & - & - \\ \hline
    \gls{ldcgan} & - & - & - & - & 4.18 & - & - & 3.45 & - \\ \hline
    \end{tabular}
    \label{table:comparisonimagefromtext}
\end{table*}

\subsection{Language generation}
\label{section:Language generation}
\glspl{gan} models have been used during the last years in \gls{nlp} tasks. The previously mentioned text-to-image field is one of the applications of \gls{gan} where natural language is involved. But there are some applications of \gls{gan} completely focused on how to produce new text using the models.

Previous methods to process natural language used the known as \textit{\gls{lstm}}~\cite{hochreiter1997long}. \gls{lstm} is capable of maintaining local relationships in space and time, this feature provides the networks the ability of process whole sentences, paragraphs and text while maintaining global coherence. In addition to \gls{lstm} the previous methods used \gls{rnn} to generate new texts~\cite{dai2015semi}.

The \gls{textgan}~\cite{zhang2016generating} uses \gls{lstm} along with \gls{cnn} to synthesize new text. The proposed method applies the \gls{gan} training methodology via the known as \textit{adversarial training}. The \gls{textgan} uses a \gls{lstm} as the G of the network and a \gls{cnn} as the D. One of the main problems of the \gls{textgan} was the highly entangled features of the network, making the interpolation of different writing styles very difficult.

The \gls{textgan} approach to language generation, suffering from the known as \textit{exposure bias}. This bias is caused by the objective function of the network, that focus on maximizing the log likelihood of the prediction. The exposure bias is visible in the inference stage, when the G generates a sequence of words iteratively predicting each word based on the previous ones. The problem comes when the prediction is based on words never seen before in the training stage. Some works were made to tackle this problem~\cite{bengio2015scheduled} but the \gls{seqgan}~\cite{yu2017seqgan} is the architecture that betters the results produced.

The G of \gls{seqgan} is trained using a stochastic policy of \gls{rl}. The \gls{rl} reward is calculated by judging a complete sentence made with the G of the model. Then, to compute the intermediate steps a Monte Carlo Search is made~\cite{browne2012survey}. The results of the \gls{seqgan} shows a huge improvement in tasks such as language generation, poem composition and music generation. In addition, the performance of the models shows certain creativity in the synthesized data.

Despite the good results of \gls{gan} in \gls{nlp} tasks during the last years, there have been developed architectures that outperform \glspl{gan} in language generation. The most successful architecture of this field is the \gls{gpt3}~\cite{floridi2020gpt}, which belongs to the GPT-n series. The \gls{gpt3} is a generator model based on the transformer~\cite{vaswani2017attention} architecture. The extraordinary results presented by the \gls{gpt3} are often very difficult to distinguish from human writing. The emergence of the \gls{gpt3} caused a lower interest in \gls{gan} models applied to \gls{nlp}. Due to the good results of transformers in \gls{nlp}, the \gls{gan} approximation to this field has been losing interest.

\subsection{Data augmentation}
\label{section:DataAugmentation}
Other field where \glspl{gan} have shown to be really useful is in data augmentation. Due to the particularities of the \gls{gan} they can be used to obtain more samples of an origin data distribution, replicating its distribution. This way, by using \glspl{gan}, the number of samples of a dataset can be multiplied.

Traditionally, data augmentation was achieved via transforming the initial data; \eg cropping, rotating, shearing, or flipping images. One of the main drawbacks of these methods is that they transform the original data by slightly changing their structure, with the usage of \glspl{gan} for data augmentation the new samples tries to synthesize new data from the original distribution. Instead of changing the samples of the dataset the generated samples of \gls{gan} are synthesized from scratch. This way, the new data is replicated by imitating the original data distribution. It should be noted that data augmentation does not necessarily replace other methods of data augmentation, it proposes an alternative that, in many cases, can be used together with other data augmentation algorithms.

For example, the \gls{dag}~\cite{tran2021data} proposes an enhanced data augmentation method for \gls{gan}, combining it with data transformation such as rotation, flipping or cropping. The \gls{dag} shows to improve the performance of data augmentation in \gls{gan} models, improving the \gls{fid} of \gls{cgan}, \gls{ssgan} and \gls{cyclegan}. The proposed architecture uses one D for each transformation of the data, but a unique G.

Data augmentation with \glspl{gan} have been used in cases where obtaining a dataset is difficult. For example, in medical applications there is usually not many information available, in this cases \glspl{gan} can make the difference. This is why during the last years \glspl{gan} have been used in medical data augmentation~\cite{frid2018synthetic, kiyasseh2020plethaugment, qi2020sag, hammami2020cycle}.

\subsection{Other domains}
\label{section:OtherDomains}
As mentioned before, due to the particularities of the \glspl{gan} they can be applied to many different fields. One of the main strengths of the machine learning is that it adapts to different situations without substantial changes in its structure. In particular, \gls{gan} can be adapted to any type of data distribution as long as there is an available dataset.

\subsubsection{GameGAN}
\label{section:gamegan}
One of the most interesting applications of \gls{gan} is the presented with the GameGAN~\cite{kim2020learning}. The main purpose of GameGAN is to generate entirely a video game using machine learning. To do so, the complete \gls{mvc} software design patterns is replicated using artificial intelligence. The proposed architecture is composed by three different modules.

The \textit{dynamics engine} is in charge of the logic of the whole system, maintaining the global coherence and updating the internal state of the game. The dynamics engine, for example, controls which actions of the game are possible (\eg eating a fruit in pac-man) and which ones are not (\eg run through a wall in pac-man). The dynamics engine is composed by an \gls{lstm} that updates the state of the game in each frame, the \gls{lstm} provides the network way to control the previous states of the game to calculate the new information of the subsequent frames. This way, the network can access to the complete history of the game, maintaining the consistency of the system.

To save the state of the game a \textit{memory module} is used. This module focus on maintain long-term consistency of the game scene. When the game is being played there are different elements of the scene that not always are visible, with the memory module these elements are consistent over the time. This memory remembers the generated static elements of the game. The memory module is implemented by using \gls{ntm}~\cite{graves2014neural}.

The third module that composes the system is the \textit{rendering engine}, it is in charge of generating a visualization of the current state of the game. This module focuses on representing the different elements of the game realistically, producing disentangled scenes. The rendering engine is composed by transposed convolution layers that are initially trained using an autoencoder architecture to warm up the system and then they train along with the rest of the modules.

The adversarial training of GameGAN has three types of discriminators. The \textit{single image discriminator} evaluates the quality of each generated frame, judging how realistic it is. The \textit{action-conditioned discriminator} determines if two consecutive frames are consistent with respect the input of the player. Finally the \textit{temporal discriminator} maintains the long-term consistency of the scene, preventing elements from appearing or disappearing randomly.

One of the basis of GameGAN is the disentangling of dynamic and static elements of the game. The static elements of a game could be, for example, walls while the dynamics elements of a game are elements such as nonplayable characters. By disentangling both types of elements, the game behavior is more interpretable for the model.

Finally, GameGAN introduces a warm-up phase where certain real frames are introduced in the network during the first epoch of the training. Then the frequency of real frames is reduced little by little until it disappears. This way the first epochs of the training, that are usually the most complex in the network, are controlled and progressively the \gls{gan} gains more control over the output. This helps the network to understand the problem.

\subsubsection{Medical imaging \glspl{gan}}
\label{section:medicalgan}
One of the most popular application of the \gls{gan} architecture is to enlarge datasets. The objective of synthesizing new data is to produce larger datasets that improve the performance of machine learning models, which are very sensible with the number of samples used in their training.

There are many fields where data augmentation can be applied, but in medical imaging to augment data have certain benefits due to the particularities of the problem. First, the medical datasets are usually small, because of the cost of obtaining the images, most of the time it is necessary to use measurement and recording machines such as radiography, magnetic resonance or ultrasound. But, in addition to the cost of obtaining these images there also exists ethical and legal problems related to the nature of the data. Most of the time, obtaining images that expose the health status of different people is impossible, which leads to even more lack of available data.

It should be noted that one of the benefits of generating data with \gls{gan} is that the new samples do not belong to any real person.

Because of all these factors, there has been lots of \gls{gan} works related to the medical imaging field~\cite{guo2020lesion, mok2018learning, uzunova2020generation, segato2020data, kossen2021synthesizing, xia2021learning}. In addition, the work of Chen et al.~\cite{chen2022generative} analyses the evolution of the field of medical data augmentation and suggests that the research in this field remains strong in the year 2021, despite that the fact that from 2019 onwards the number of published works have been the same.

\subsubsection{\glspl{gan} in agriculture}
\label{section:agriculturegan}
Similar to the medical imaging field, obtaining images to train the computer vision models of agricultural image analysis is not an easy task. These models benefit from having large-scale balanced datasets but the cost of obtaining high quality labelled data makes the data augmentation a crucial task in these datasets.

Many different \gls{gan} models have been applied to agricultural data, such as \cite{li2022fwdgan, xu2021style, jin2022grapegan}. These works aim to generate new images of plant with different diseases, augmenting the number of samples by using \gls{gan}.

In these cases the use of \gls{gan} improves the results of the machine learning models by enlarging the number of available data. The agricultural images have different particularities that make the analysis of them a difficult task. For example the biological variability between two samples of the same specie makes crucial to have many different samples to learn all the modes of the data. In particular, the same leaf of a fruit can drastically differ from one individual to another.

Other important factor is that the labelling of the data can be very costly, specially for specific applications such as the disease detection of certain plant, e.g. tomato leaf~\cite{li2022fwdgan}.

In addition, the environment where the images are taken, most of the time in crops, can lead to many variance in the images, such as lighting changes or object occlusion.

\subsubsection{Drug discovery using \glspl{gan}}
\label{section:druggan}
The process of discovering and designing new drugs has recently been impulsed by the field of Deep Learning~\cite{jing2018deep, dana2018deep}. In particular, \glspl{gan} are an useful technique to synthesize new useful samples of data. In the drug environment, the \gls{gan} architecture can process the drug compound using graphs or \gls{smiles}, to then generate synthetic samples of drugs.

Due to the flexibility that \glspl{ann} have in terms of operating with different data types, it is possible to use the same architectures in different fields. In this case the overall \gls{gan} design can be adapted to molecular data, being able  to transfer the same principles of the image generation to new data types.

The research followed by Kadurin et al.~\cite{kadurin2017cornucopia, kadurin2017drugan} generates new drug compounds for anticancer therapy, using biological and chemical datasets. In particular, in \cite{kadurin2017drugan} it is used an Adversarial Autoencoder that uses molecular fingerprints as inputs of the network. By using this architecture the researches are able of define the desired properties of the synthesized drugs. Some of the new synthetic drugs discovered by the Deep Learning architecture corresponded with previous known anticancer drugs. This led the researches to suggest that the remaining unknown drugs generated by the \gls{gan} could be used to further study their properties.

The work presented in \cite{padalkar2021drug} proposes the generation of new drugs combining \glspl{gan} with reinforcement learning techniques. In particular, the proposed G takes as input a random latent space and process it with \gls{rnn} to produce a sequence of drug by using \gls{smiles} representation. The D on its side uses a 1 dimensional \gls{cnn} to distinguish the real data from the synthesized one. The results of the paper suggest that the new drugs discovered were unique and diverse. This may alleviate the first phases of drug development, which are very expensive in terms of time.

The \gls{fldisco} architecture~\cite{manu2021fl} aims to combine the generation potential of \gls{gan} with the processing of molecules using graphs of the Graph Neural Networks while maintaining the privacy of the data using Federated Learning~\cite{konevcny2016federated}. By using graph representation of the molecules instead of \gls{smiles} as previous works, the represented samples have more realistic structures, maintaining structural relationships of the connected atoms of the molecules. The Federated Learning framework is based on using different clients to train a specific neural network model, each client has its respective portion of the data, which uses to train the network. This way each client knows a portion of the data and uses it to update the central model, but it maintains the privacy due to the fact that the clients are not able to communicate with each other. The results of this research show progress in terms of novelty and diversity of the synthesized drugs respect previous works.

\section{Discussion}
\label{section:Discussion}
Since their introduction in 2014 \glspl{gan} have been the most important generative architecture in computer vision. The results provided by the developed \glspl{gan} were notoriously better than previous architectures, such as Variational Autoencoders. This leaded to a constant improvement of the model, solving problems like stabilization or mode collapse.

With the introduction of the Diffusion models~\cite{dhariwal2021diffusion, ho2020denoising, song2019generative}, the results of \glspl{gan} have been surpassed by this new models solving some of its most important problems. Some aspects in which diffusion models outperform \glspl{gan} are better stability, they do not suffer from mode collapse and they provide more diverse results. This is mainly caused because of the fact that they are likelihood-based~\cite{croitoru2022diffusion}. Despite the better results of diffusion models they still have shortcomings in some aspects such as the cost of synthesizing new samples, which makes them difficult to being applied in real-time problems.

In \cite{saharia2022palette} it was developed a diffusion model to perform an image-to-image translation. The results showed in this research show that their solution outperforms \glspl{gan} without special attention to the hyper-parameter tunning or any kind of sophisticated technique or loss function. Moreover this research shows the great stability of the diffusion model architecture.

Despite the fact that Diffusion models are a novel architecture with not many works published, it is a very potential architecture to surpass \gls{gan} results in a near future. At present, there are not enough results or applications of diffusion models to data generation, but the potential of this new architecture could lead to a significant improvement in the results of data synthesis. We consider that this models could replace \glspl{gan} because of their stability and not needing fine-tunning in their hyperparameters.

Other new architectures have been used to enhance the results of \glspl{gan}, such as transformers, to improve their results. Transformer architecture is a time-series-based architecture that adopts the self-attention layers~\cite{vaswani2017attention} making possible to design larger models. Transformers have been used as the base neural model of the G and D of the \gls{gan} architecture, improving the performance of the model.

The TransGAN~\cite{jiang2021transgan} presents a \gls{gan} architecture free of convolutions that makes possible to generate high resolution images by using transformer in both G and D of the \gls{gan}. The results of the article shows an improved results respect to the \gls{is} and \gls{fid} on CIFAR-10 dataset~\cite{krizhevsky2009learning}.

Another work that showcases the interaction between \glspl{gan} and transformers is the one presented in \cite{lv2022improved}. This work uses the generative model to predict pedestrian paths, using the memory that the transformer architecture has. In this sense, the \gls{gan} makes possible to train the network to predict future paths of pedestrians, while the transformer provides the memory to process an historical sequence of the latest movements.

\section{Conclusion}
\label{section:Conclusion}
This report summarizes the recent progress of \glspl{gan}, going from the basic principles in which \gls{gan} are sustained to the most innovative architectures of the last years. In addition, the different problems that \glspl{gan} can suffer are categorized and the most common evaluation metrics are explained and discussed.

Respect the recent progress in the field, a taxonomy for the \gls{gan} variants is proposed. The researches are divided in two groups, one with the \glspl{gan} that focus in architecture optimization and the other with the \glspl{gan} that focus in objective function optimization. Despite being two separate groups of variants, it should be noted that the different researches benefit from the progress of the rest. These ecosystem where there are various approaches for \gls{gan} development is connected with the main problems that are reviewed in this survey, since normally each research focus in trying to solve a certain problematic of previous researches.

Finally the different application of the \glspl{gan} during the last years are summarized. The different applications of \gls{gan} are influenced by the development of the field, its impact in the society and in the industry. We conclude with a comparison between the different architectures performance to provide a quantitative view of the evolution of \glspl{gan}.


\newpage

\bibliography{references}

\end{document}